\documentclass[final]{cvpr}

\usepackage{times}
\usepackage{epsfig}
\usepackage{graphicx}
\usepackage{amsmath}
\usepackage{amssymb}
\usepackage{multicol}
\usepackage{multirow}
\usepackage{xr}
\usepackage{makecell}
\usepackage{xcolor}
\usepackage[ruled,vlined]{algorithm2e}
\usepackage{algorithmic}
\usepackage{pifont}

\makeatletter
\newcommand*{\addFileDependency}[1]{
  \typeout{(#1)}
  \@addtofilelist{#1}
  \IfFileExists{#1}{}{\typeout{No file #1.}}
}
\makeatother

\newcommand*{\myexternaldocument}[1]{%
    \externaldocument{#1}%
    \addFileDependency{#1.tex}%
    \addFileDependency{#1.aux}%
}
\newcommand{\cmark}{\ding{51}}
\newcommand{\xmark}{\ding{55}}
\newcommand{\main}[1]{\textcolor{black}{#1}}

\myexternaldocument{supp}

\usepackage[pagebackref=true,breaklinks=true,colorlinks,bookmarks=false]{hyperref}

\def\cvprPaperID{6819} 
\def\confYear{CVPR 2021}
\def\FewC{Scarce-Class}
\def\FewI{Scarce-Image}
\def\benchmark{LRDS}

\begin{document}
\title{Unlocking the Full Potential of Small Data with Diverse Supervision}

\author{
$\text{Ziqi Pang*}^1 \text{\qquad Zhiyuan Hu*}^2 \text{\qquad Pavel Tokmakov}^3$\\
$\text{\qquad Yu-Xiong Wang}^4 \text{\qquad Martial Hebert}^5$\\
$^1\text{TuSimple \qquad} ^2\text{UCSD \qquad} ^3\text{Toyota Research Institute \qquad} ^4\text{UIUC}\qquad ^5\text{CMU}$\\
\tt\small ziqi.pang@tusimple.ai, z8hu@ucsd.edu, pavel.tokmakov@tri.global,\\
\tt\small yxw@illinois.edu, hebert@cs.cmu.edu
}

\maketitle

\begin{abstract}
Virtually all of deep learning literature relies on the assumption of large amounts of available training data. Indeed, even the majority of few-shot learning methods rely on a large set of ``base classes'' for pre-training. This assumption, however, does not always hold. For some tasks, annotating a large number of classes can be infeasible, and even collecting the images themselves can be a challenge in some scenarios. 
In this paper, we study this problem and call it ``Small Data'' setting, in contrast to ``Big Data.'' To unlock the full potential of small data, we propose to augment the models with annotations for other related tasks, thus increasing their generalization abilities. In particular, we use the richly annotated scene parsing dataset ADE20K to construct our realistic Long-tail Recognition with Diverse Supervision (LRDS) benchmark, by splitting the object categories into head and tail based on their distribution. Following the standard few-shot learning protocol, we use the head classes for representation learning and the tail classes for evaluation. Moreover, we further subsample the head categories and images to generate two novel settings which we call ``Scarce-Class'' and ``Scarce-Image,'' respectively corresponding to the shortage of training classes and images. Finally, we analyze the effect of applying various additional supervision sources under the proposed settings. Our experiments demonstrate that densely labeling a small set of images can indeed  largely remedy the small data constraints. Our code and benchmark are available at \url{https://github.com/BinahHu/ADE-FewShot}.
\end{abstract}

\vspace{-10px}
\section{Introduction}
\label{sec::intro}
The availability of big data is the cornerstone of deep learning research. From early successes on image classification that were enabled by ImageNet~\cite{imagenet}, to more recent progress in object detection driven by the large-scale COCO dataset~\cite{lin2014microsoft}, thousands of carefully curated and accurately annotated images are essential to outperform the classical, heuristic-based approaches. Indeed, even the latest unsupervised learning methods rely on availability of millions of images from the target domain~\cite{he2020momentum}. This requirement, however, cannot always be fulfilled. 

\begin{figure}
	\centering
	\includegraphics[width=0.95\columnwidth]{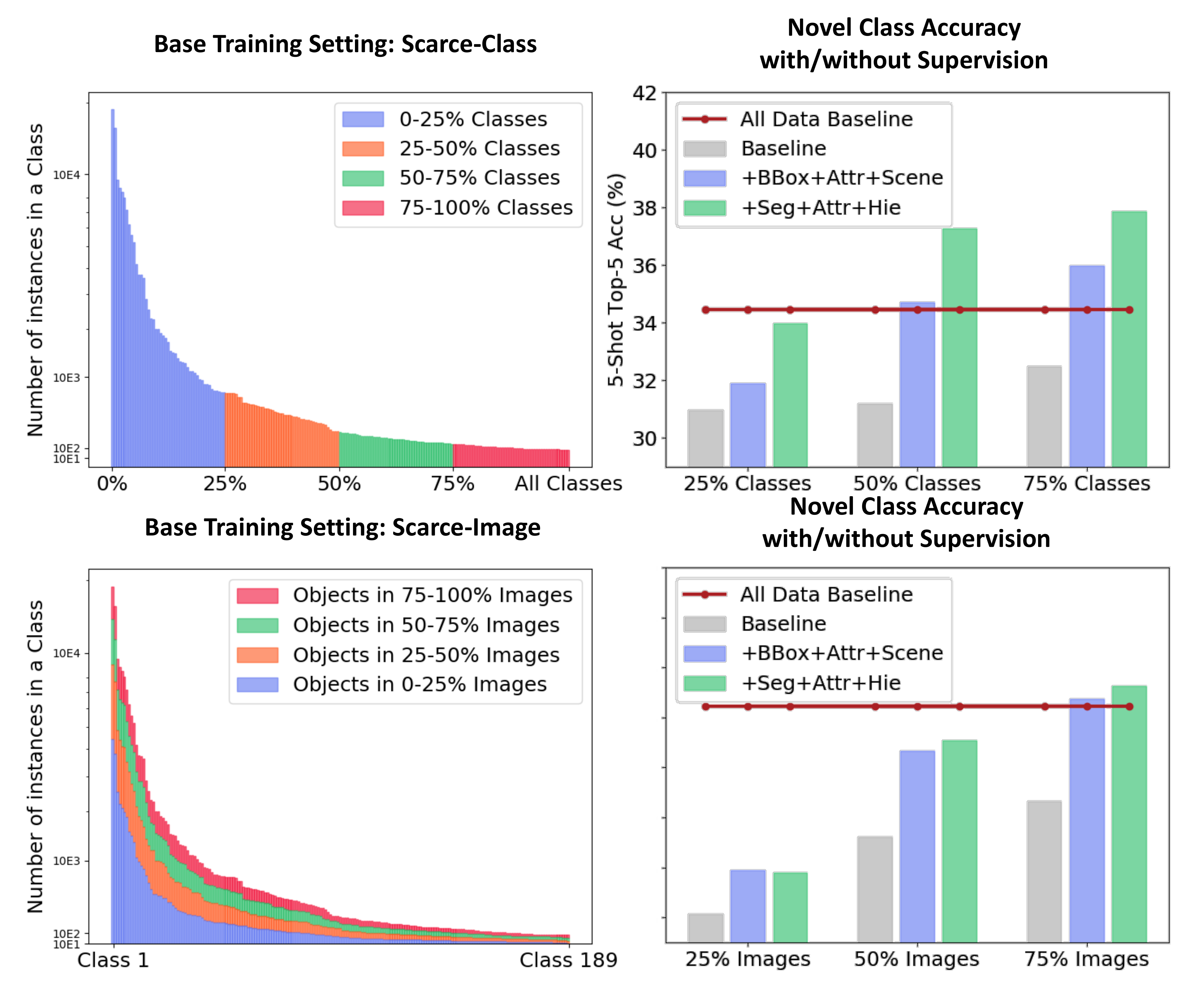}
	\caption{``Small Data'' settings and the effect of using diverse supervision. \textbf{Left:} We visualize the category-level instance distribution of ``Small-Data'' regimes. Specifically, \FewC~picks the frequent classes and \FewI~randomly removes a portion of images. \textbf{Right:} Baseline represents the models trained with classification labels only. By incorporating additional supervision, we achieve competitive performance with less data. The types of supervision used: Bounding Box (BBox), Attributes (Attr), Scene Labels (Scene), Segmentation (Seg), and Class Hierarchy (Hie).}
	\vspace{-15px}
	\label{fig::cover}
\end{figure}

First, in many domains the category distribution is highly skewed, with a few categories covering most of the data (so-called head of the distribution), and the rest having only several examples per category (so-called tail)~\cite{lvis,gu2018ava}. This setting limits the performance of the methods that rely on a large set of head categories for representation learning. Second, the collection of raw images themselves can be difficult. For instance, in face recognition privacy concerns can limit the data collection~\cite{erkin2009privacy, ma2019lightweight}, and in medical imaging the amount of examples is constrained by the high cost of the image acquisition devices~\cite{zhao2019data, ouyang2020self}. 

The problem of learning under data sparsity constraints has been mostly studied in the few-shot learning setting. Earlier works used toy datasets, like Omniglot~\cite{lake2015human}, or mini-ImageNet~\cite{vinyals2016matching}. These benchmarks, however, were not realistic both in terms of content, featuring small, object-centric images, and in terms of data distribution, with categories being artificially balanced and randomly split into head and tail. The issue of content realism was partially addressed in the full ImageNet~\cite{hariharan2017low}, and more recently in MetaDataset~\cite{triantafillou2020meta}. In addition to content realism, the latter benchmark attempted to achieve distribution realism as well by avoiding artificially balancing the categories. Both of these datasets, however, still remain object-centric, and ignore the role of context in recognition. More importantly, by virtue of their size, they have set a new standard of pre-traing representations on a large collection of head classes, which, as we have described above, is not always possible in practice. 

In this work, we address the limitation of existing few-shot learning benchmarks by introducing a new, realistic dataset --  Long-tail Recognition with Diverse Supervision (\benchmark). Instead of collecting such a dataset from scratch, we re-purpose the existing ADE20K dataset~\cite{ADE20K}, originally for scene parsing. This dataset was collected by labeling all the objects in a diverse set of images with an open vocabulary, which resulted in a natural long-tail distribution of categories (see Figure~\ref{fig::benchmark}, left). Moreover, since the objects were labeled in context, we are able to sample them together with a significant portion of the background, removing the object-centric bias of existing benchmarks (see Figure~\ref{fig::ade_convert}).

To better emulate the real-world challenges of learning from small data, we propose two new evaluation settings in \benchmark: \textbf{\FewC} and \textbf{\FewI}, which are illustrated in the left column of Figure~\ref{fig::cover}. The former emulates the heavy skew towards the tail categories by removing the least frequent classes from the head, whereas the latter targets the difficulty of image collection by randomly subsampling images in the training set of the head categories. As can be seen in the right column of Figure~\ref{fig::cover}, both these settings lead to a significant performance degradation on the tail classes with respect to the baseline trained on the full head set, emphasizing the difficulty of the problem. 


To unlock the full potential of small data, we explore additional sources of supervision that can be helpful to enrich the representations without the need to collect new images or labels for the main task (such additional supervision can be a lot easier to obtain -- one can always annotate bounding boxes in existing images, whereas collecting new images may be infeasible). These experiments are enabled by the wide variety of annotations available in ADE20K. In particular, we study the effect of localization supervision in the form of object masks and bounding boxes, background segmentation, scene-level labels, and object part annotations, as well as semantic supervision in the form of attributes and class hierarchy (see Figure~\ref{fig::benchmark}, right). In addition, we show that these forms of supervision can be combined to further improve the performance. As illustrated in the right column of Figure~\ref{fig::cover}, diverse supervision can cover up the degradation from lacking training data to some extent.

To sum up, our contributions are three-fold: (1) We propose a novel benchmark for evaluating long-tail and few-shot learning in a realistic setting. It is based on the ADE20K dataset for scene parsing and features a natural long-tail distribution of object categories; (2) We introduce two new evaluation settings -- \FewC~and \FewI, which emulate the realistic issues of class and image scarcity; (3) We demonstrate that incorporating diverse supervision can help unlock the full potential of small data by enriching the learned representations, thus increasing their generalization abilities.

\vspace{-5px}
\section{Related Work}
\label{sec::related_work}
\vspace{-2px}
{\bf Related visual benchmarks.} Our proposed \benchmark\ benchmark aims to facilitate further investigation on few-shot learning (FSL) and long-tail learning (LTL). Most of the existing benchmarks for FSL and LTL are constructed from balanced datasets like ImageNet~\cite{imagenet}. Then the data distribution is {\em manually} modified to simulate the data imbalance and/or scarcity, by splitting the classes into head and tail and subsampling the tail classes. Representative benchmarks include mini-ImageNet~\cite{vinyals2016matching}, tiered-ImageNet~\cite{ren2018meta}, and CUB~\cite{CUB} for FSL, Places-LT~\cite{liu2019large} and ImageNet-LT~\cite{liu2019large} for LTL, and BSCD-FSL~\cite{guo2020broader}, MetaDataset~\cite{triantafillou2020meta}, and~\cite{tseng2020cross} that involve multiple datasets for cross-domain FSL. Despite the progress made possible by these benchmarks, they are still not practical -- their random splitting and subsampling strategies fail to properly reflect how the visual concepts are naturally organized. Also, either common or rare categories have inherent properties (\eg, scale, context, or super-category), but these important properties are lost during such process. Our \benchmark\ instead captures the real-world frequency from ADE20K~\cite{ADE20K}, and is thus more natural and realistic for both FSL and LTL. In addition, no existing benchmark investigates the practical settings with \FewC\ and \FewI.

\vspace{-2px}
Most relevant to our benchmark, iNaturalist~\cite{wertheimer2019few} is also a realistic long-tail dataset~\cite{inaturalist}. However, it only contains animals, while our \benchmark\ benchmark has 482 categories covering diverse concepts of living and non-living, objects indoor and outdoor. Moreover, \benchmark~also exhibits a more severe head/tail imbalance, with the frequencies ranging from 15 to more than 20,000, which poses serious challenge to existing long-tail techniques. LVIS~\cite{lvis} is a recent dataset also capturing a natural long-tail distribution. However, LVIS does not have additional annotation information, and is thus not suitable for our exploration. By contrast, our \benchmark~contains rich annotations reflecting the complexity of the visual world. 

\vspace{-2px}
Multi-task learning (MTL) is also related. To the best of our knowledge, current MTL benchmarks~\cite{standley2020which, nyud, pascal} focus on pixel-level RGBD tasks with relatively abundant data. Our benchmark is the first to incorporate the few-shot classification problem into the context of MTL.

\vspace{-2px}
{\bf Few-shot learning.} FSL is a classical problem of recognizing novel objects from few training examples~\cite{thrun1996learning,mishra2018a, oreshkin2018tadam, lee2019meta, wu2018improving, qiao2018few, rusu2019meta}. Some representative approaches include metric learning~\cite{ koch2015siamese,vinyals2016matching,protonet,yang2018learning}, meta-learning ~\cite{maml,ravi2016optimization,wang2016learningfrom}, and their combination~\cite{triantafillou2020meta}. However, some most recent work~\cite{gidaris2018dynamic, closerlook, tian2020rethinking} shows that the performance of these complex models can be matched by simple representation learning on base classes and classifier fine-tuning on novel classes. Consistent with and further supportive to this observation, in the paper we mainly focus on the incorporation of supervision with the linear classifier baseline~\cite{closerlook, tian2020rethinking}, though we also provide results for representative FSL methods on our \benchmark~benchmark.

\vspace{-2px}
Some FSL work improves feature learning with external cues, which is the most relevant to ours.~\cite{gidaris2019boosting,dvornik2019diversity, su2020when} use self-supervised tasks. ~\cite{wertheimer2019few, lin2020few} utilize localization information. Many others explore semantic information:~\cite{li2019large} uses the hierarchical structure of concepts, ~\cite{tokmakov2018learning, huang2020attribute} use attributes, and~\cite{schwartz2019baby, xing2019adaptive, mu2020shaping} exploit more extensive semantic information. However, ours is the first that systematically investigates extensive and diverse sources of supervision and further addresses their combination, thus pushing the state of the art in this research direction.

\vspace{-2px}
{\bf Multi-task learning.} MTL aims at learning a single model to handle multiple tasks~\cite{argyriou2007multi}. Most of the work seeks to jointly maximize the performance of each individual task. Existing approaches can be mainly grouped into two categories: balancing the weights of loss functions~\cite{sener2018multi, kendall2018multi, chen2018gradnorm} and introducing better feature sharing and transferring mechanisms~\cite{mtinet, strezoski2019many,bragman2019stochastic, maninis2019attentive, rebuffi2017learning, rebuffi2018efficient, xu2018padnet, kokkinos2017ubernet}. In contrast to their objective of boosting all the involved tasks, we combine multiple sources of supervision to improve the representation learning for classification. In our experiments, we show that a simple sequential and multi-task training approach achieves competitive performance under this objective.

{\bf More broadly}, our work is related to the general investigation of learning with varying amount of data and annotation, including fully-supervised learning~\cite{imagenet, krizhevsky2017imagenet}, semi-supervised learning, and self-supervised learning. While such work~\cite{sun2017revisiting,goyal2019scaling} is still in the regime of big data with weak or no annotation, here we in particular focus on the ``opposite'' direction: ``Small Data'' with rich supervision.

\vspace{-5px}
\section{The \benchmark\ Benchmark}
\label{sec::benchmark}
\vspace{-5px}
In this section, we construct a new benchmark for recognizing rare categories in the wild. The benchmark features a realistic and diverse class distribution, with objects captured in context. Additionally, we use this benchmark to study \FewC~and \FewI~regimes, and investigate to what extent can diverse supervision help alleviate the issues caused by data insufficiency. Below, we describe the steps we take to satisfy these objectives.

\begin{figure}
  \includegraphics[width=1.0\columnwidth]{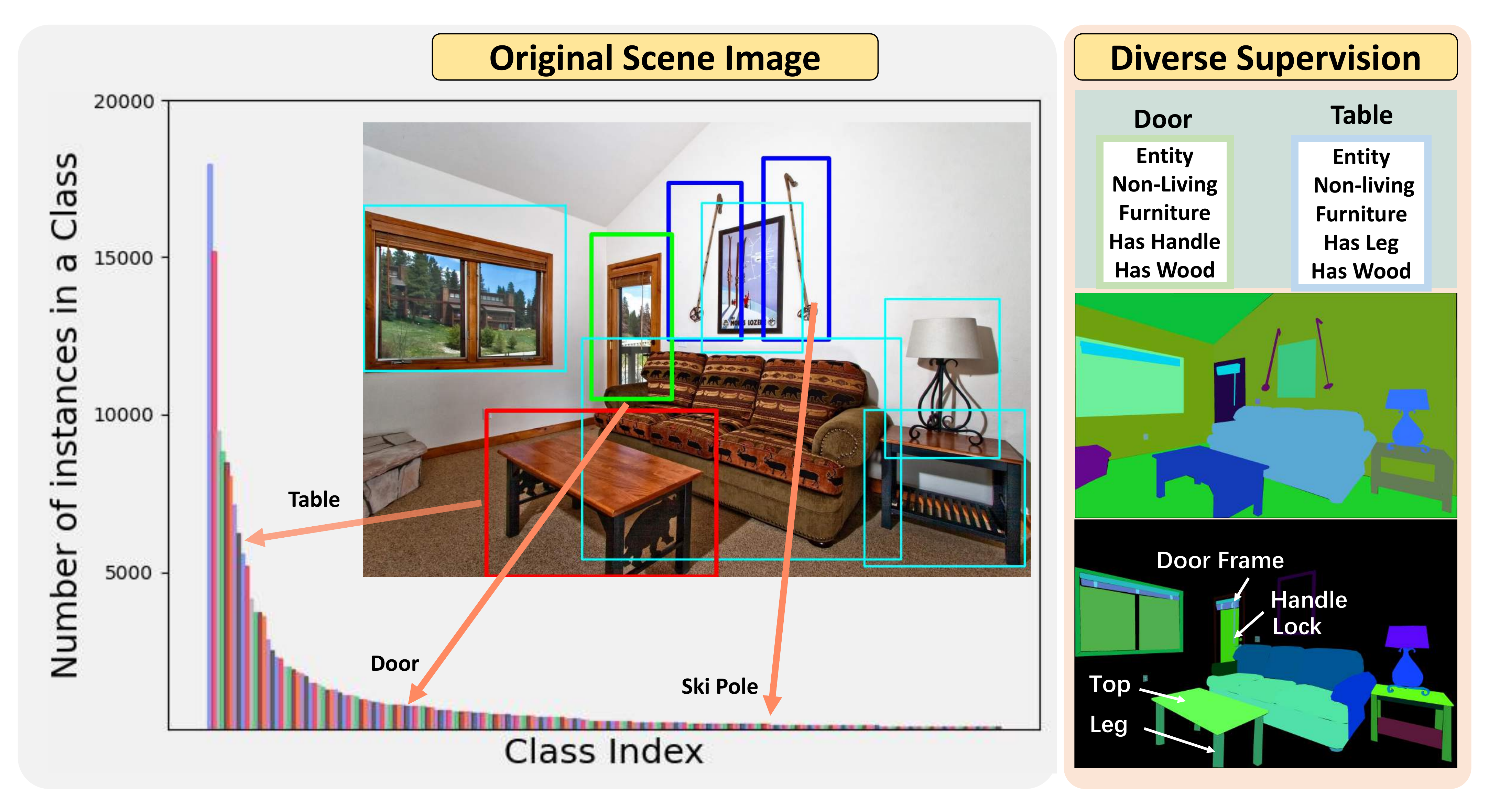}
  \caption{Our \benchmark\ benchmark. \textbf{Left:} \benchmark\ is constructed by turning the scene parsing dataset ADE20K into a long-tailed image classification benchmark. \textbf{Right:} We present three types of supervision for example: Attributes, Segmentation, and Parts, which are used for learning rich features in the ``Small Data'' regime. }
  \label{fig::benchmark}
  \vspace{-15px}
\end{figure}

\vspace{-5px}
\subsection{Data Selection and Splitting}
\label{subsec::cat_selection}
We choose the ADE20K~\cite{ADE20K} dataset as the basis for constructing our few-shot learning benchmark, due to its diversity and richness of the vocabulary. ADE20K has more than 3,000 classes, covering objects, object parts, and background. It also exhibits a highly imbalanced category distribution shown in the left column of Figure~\ref{fig::benchmark}. Below we summarize the filtering steps to select the categories for our benchmark.

First, we manually split the classes into objects, parts, and stuff. Since we want to focus on object classification, only the corresponding categories are included in our label set, resulting in 1,971 classes. The part and stuff labels are instead used as additional sources of supervision. 

Second, we further filter the object categories based on their frequency. In particular, we only keep the classes with at least 15 instances in the dataset, resulting in 482 categories. While this excludes the most challenging classes from the benchmark, we argue that including them would introduce significant noise in the evaluation. Indeed, measuring performance on 2-3 images is dominated by noise and is not informative of the model's recognition ability.

Finally, to mimic the typical setup in the few-shot classification benchmarks, we split the categories into base and novel subsets, where the former is used for representation learning, and the latter for evaluation. Instead of splitting the categories at random, we follow the natural distribution of objects in the world, and select the classes that have more than 100 instances in the dataset as base and the remaining ones as novel. This results in a realistic few-shot learning benchmark to date, where the natural regularities between frequent and infrequent categories are captured.

Overall, our dataset contains 189 base and 293 novel categories. For each of the categories in the base set, 1/6 of the data is held out for validation. In the novel set, we randomly select 5 instances in each category for training and the rest are used for evaluation. We further divide the novel set into 100 novel-val and 193 novel-test categories at random, where the former is used for hyper-parameter selection, and the latter for reporting the final performance.

\subsection{Adapting ADE20K for Classification}
\label{subsec::convert}
\begin{figure}
    \centering
    \includegraphics[width=0.7\columnwidth]{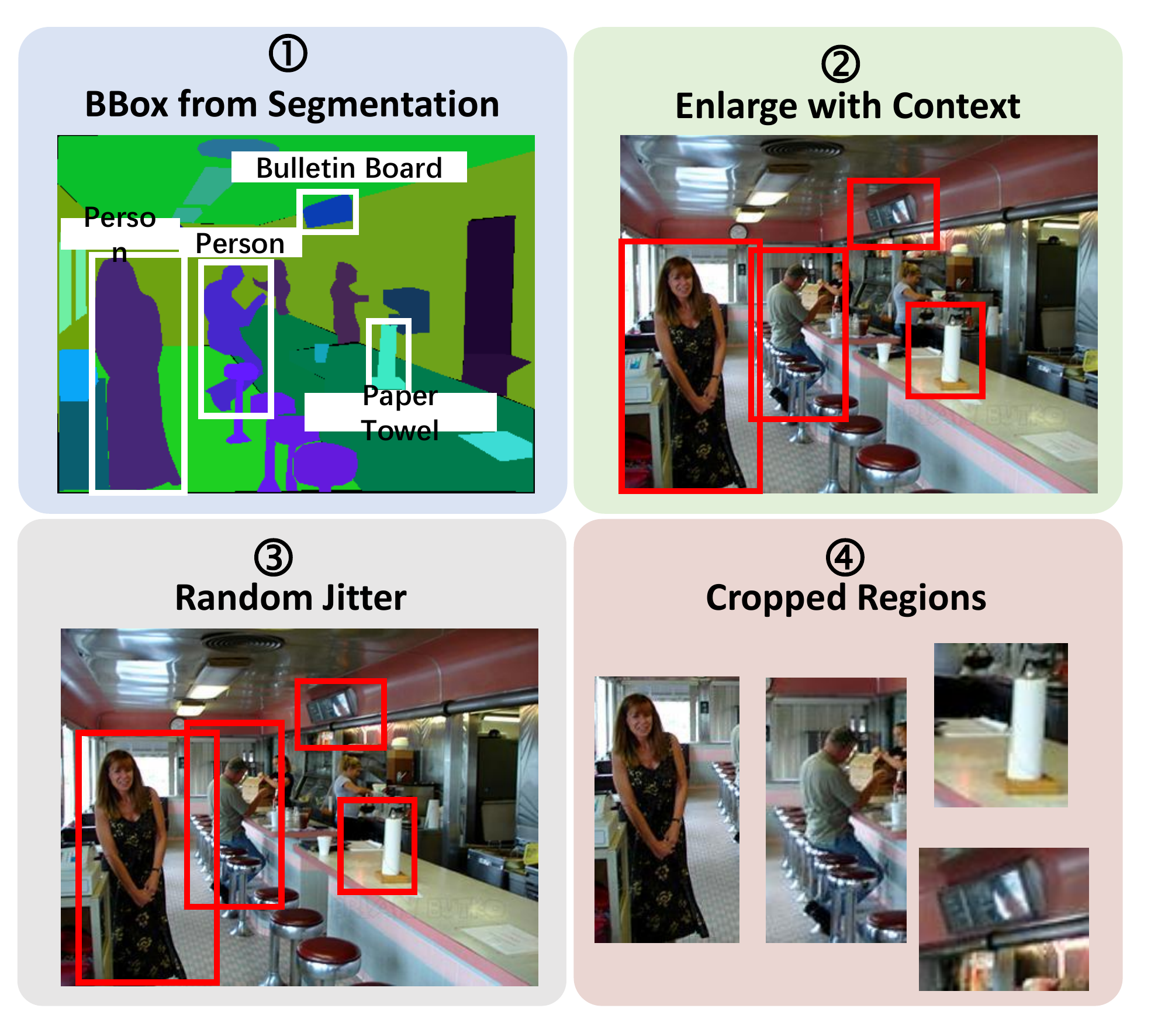}
    \caption{Converting ADE20K into an image classification benchmark. With the original segments annotated in ADE20K, we first crop out every object using tight bounding boxes. Then we imitate realistic data distribution by enlarging the boxes with context and applying random jitter to avoid center bias.}
    \label{fig::ade_convert}
    \vspace{-15px}
\end{figure}
We convert the original ADE20K designed for scene parsing into a classification dataset to serve our needs. Since ADE20K provides instance-level masks for the objects, it can be easily transformed by cropping the regions around the masks and treating them as independent images. However, using tight bounding boxes is an unrealistic data distribution, since objects typically appear in context. Therefore, we simulate a more realistic distribution by box enlargement and random jitter. In particular, we compute the average context ratio (area of context divided by area of tight bounding box) in the ImageNet~\cite{imagenet} dataset and expand the original bounding boxes accordingly. We then apply a random shift to the box to avoid center bias (see Figure~\ref{fig::ade_convert}).

\subsection{Defining the Small Data Regimes}
\label{subsec::small_regime}
To experiment under the ``Small Data'' settings, we propose two new evaluation regimes, \FewC~and \FewI, to simulate the corresponding real-world constraints. For the former, we sort the base categories according to the number of instances, and remove the lower 25\%, 50\%, and 75\% for varying degrees of class sparsity. This indeed corresponds to highly-skewed long-tailed distributions in the real world when increasingly fewer categories occupy a larger fraction of the data. For the latter, we simply remove the images from the training set of the base categories at random using the same ratios as for the \FewC~regime. This approach simulates  scenarios in which the images are hard to obtain, while preserving the natural, long-tailed category distribution. 

Importantly, these two settings have a different effect on the overall size of the training set. Recall that in \FewC~the least frequent categories are removed first, which constitute only a small fraction of the dataset (see Figure~\ref{fig::cover}, top left). Removing images, on the other hand, has a more monotonous effect on the dataset size, as shown in the bottom left part of Figure~\ref{fig::cover}. As a results, the two settings have a different effect on the model's performance, with \FewI~being harder in the most constrained scenario, when only 25\% of the data is preserved (see Figure~\ref{fig::cover}, right).

\subsection{Collecting Diverse Supervision}
\label{subsec::sup_collection}

Evaluating the effect of diverse supervision on the model’s few-shot classification ability is one of the main goals of this work. To this end, we accumulate all the labels provided in ADE20K, which include localization supervision for the object categories (both in the form of masks and bounding boxes), object part annotations, stuff category segmentation, as well as scene labels. In addition, inspired by~\cite{tokmakov2018learning}, we collect category-level attribute annotations by inheriting the same attribute set defined on ImageNet~\cite{imagenet}, and manually assigning attributes for each base category. Finally, we collect class hierarchy labels by extracting the WordNet Tree provided by ADE20K. Notice that {\em no additional annotations are used for the novel categories}.

Combining all these diverse forms of supervision in a single image classification framework is non-trivial. In the next section we discuss our experimental setup as well as evaluation protocol.






\vspace{-5px}
\section{Experimental Protocol}
\label{subsec::setup}
\begin{figure*}
    \centering
    \includegraphics[width=2.0\columnwidth]{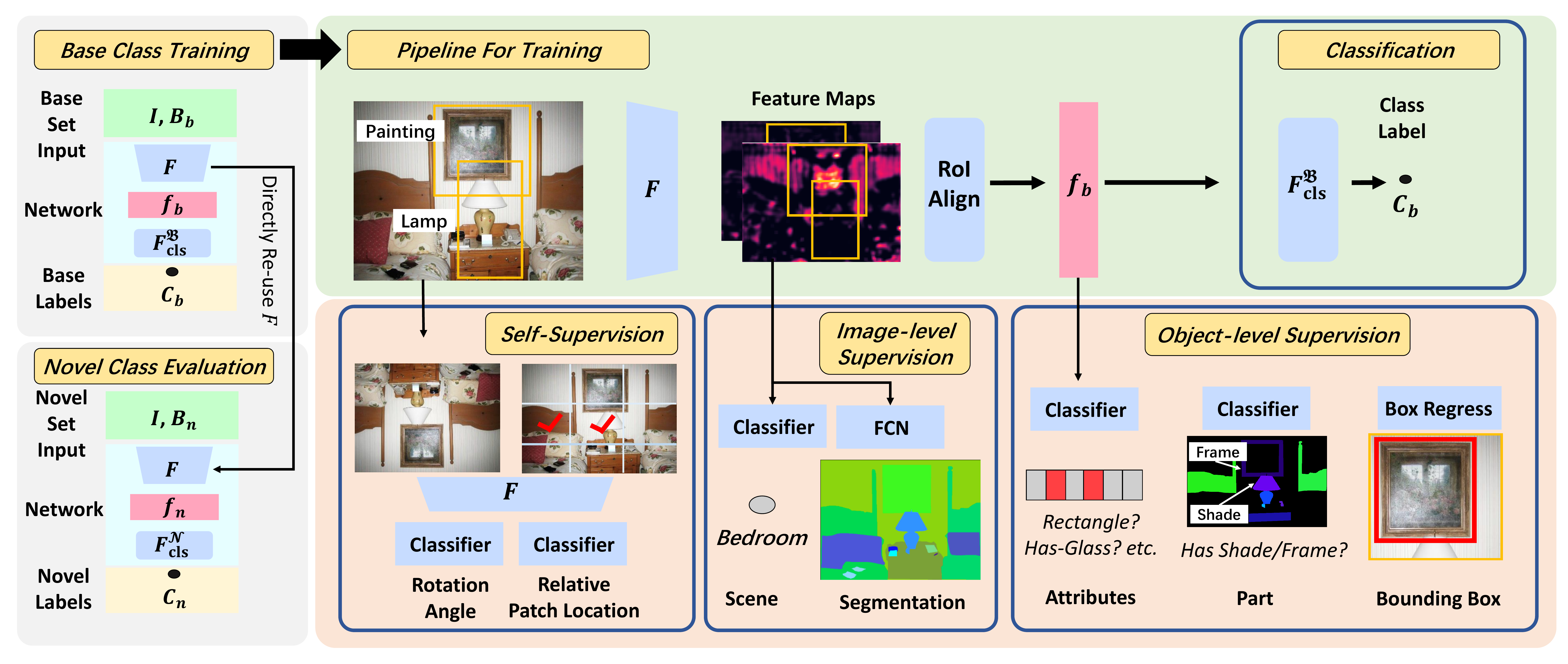}
    \caption{Pipeline for training and evaluation with diverse supervision. \textbf{Left:} We learn the feature extractor $F$ on base classes and evaluate it on novel classes for few-shot learning. \textbf{Upper-Right:} The model is based on the classification branch of Faster R-CNN~\cite{FasterRCNN}. \textbf{Lower-Right:} We show some representative examples of introducing supervision at different levels. Self-supervision modifies the input image, image-level supervision operates on the entire feature map, and object-level supervision focuses on the feature vectors of the objects.}
    \label{fig::model}
    \vspace{-15px}
\end{figure*}
The pipeline and setup of our experiments are demonstrated in Figure~\ref{fig::model}, where the standard few-shot learning protocol in~\cite{closerlook} is adopted. We first learn the feature extractor $F$, such as a ResNet-18~\cite{resnet}, on the training set of base classes. Then $F$ is frozen and evaluated on novel classes for few-shot learning. The accuracy on the test set of novel classes indicates the generalization of the learned feature. The evaluation focuses on both 1-shot and 5-shot settings. Note that, different from~\cite{closerlook}, we perform classification on the full set of classes (100-way for the validation set and 193-way for the test set).

During both training and evaluation, a model takes as input a region $B$ in a scene image $I$, and outputs the category $C$ of the object in that area. Therefore, our backbone is identical to the classification branch of Faster R-CNN~\cite{FasterRCNN}, which performs object classification on the input image regions. The scene image $I$ is propagated through $F$, producing a feature map. Then the feature map of the region $B$ is fed into an RoI-Align layer~\cite{maskrcnn}, whose output is the feature vector $f$ of the region. Finally, the model predicts the category $C$ with a classifier, denoted as $F_{\mathtt{cls}}^{\mathcal{B}}$ and  $F_{\mathtt{cls}}^{\mathcal{N}}$ for base and novel classes, respectively. More formally, assume that the image $I$ consists of object set $\mathcal{O}_I$; then the classification loss for the base objects is as Eq.~\eqref{eq::cls}, where $\mathcal{L}_{\mathtt{CE}}$ and $\mathtt{RA}$ denote the cross-entropy loss and the RoI-Align operation:
\begin{equation}
\begin{aligned}
\mathcal{L}_{\mathtt{cls}}(I)=\sum\limits_{o \in \mathcal{O}_I}\mathcal{L}_{\mathtt{CE}}(F_{\mathtt{cls}}^{\mathcal{B}}\circ \mathtt{RA}(F(I), B_o), C_o).
\end{aligned}
\label{eq::cls}
\end{equation}

When combining different types of supervision, each type of supervision $s$ has its own task head $F_s$, supervision label $C_s$, and task-specific metric $\mathcal{M}_s$. Taking the \textit{Scene Label} supervision as an example, we have a classifier, scene label, and cross-entropy loss.

The object-level supervision applies $F_s$ on top of the feature vectors of objects. This set of supervision includes Attributes, Class Hierarchy, Parts, and Bounding Box. For an arbitrary supervision $s$, the loss on image $I$ is computed as:
\begin{equation}
\begin{aligned}
\mathcal{L}_s(I)=\sum\limits_{o\in\mathcal{O}_I}\mathcal{M}_s(F_{s}\circ \mathtt{RA}(F(I), B_o), C_{o,s}).
\end{aligned}
\label{eq::obj_level}
\end{equation}

For the type of image-level supervision which is relevant to the entire image, such as Scene Label and Segmentation, we apply $F_s$ on top of the feature map $F(I)$. Their loss functions are in the form of
\begin{equation}
\begin{aligned}
\mathcal{L}_s(I)=\mathcal{M}_s(F_s\circ F(I), C_{s}).
\end{aligned}
\end{equation}

In addition, we further consider using pretext tasks as self-supervision~\cite{gidaris2019boosting,dvornik2019diversity}. This type of supervision defines an editing operation $E$ on the images, and predicts labels corresponding to $E$. The losses for self-supervision $s$ are
\begin{equation}
\begin{aligned}
\mathcal{L}_s(I)=\mathcal{M}_s(F_s\circ F\circ E_s(I), C_s).
\end{aligned}
\end{equation}

When multiple types of supervision are combined, the final loss function $\mathcal{L}$ becomes the weighted sum of each individual loss term.
The combination weights are hyper-parameters that balance different supervision and are selected on the validation set. We elaborate on this and other implementation details in the appendix.

\vspace{-5px}
\section{Experimental Evaluation}
\label{sec::experiment}
We begin by evaluating state-of-the-art approaches for few-shot learning on our benchmark in Section~\ref{subsec::meth}. Then we use the full training set of the base classes to study the effect of using additional sources of supervision in Section~\ref{subsec::individual_sup} and lowering the cost of supervision in Section~\ref{subsec::small_sup}. In advance, we explore the optimal ways of combining them in Section~\ref{subsec::mult_sup}. Finally, we evaluate the best combinations in the very challenging \FewC~and \FewI~regimes in Section~\ref{sec::less_data} (more analysis is reported in the appendix).

\subsection{Benchmarking Few-shot Learning Methods}
\label{subsec::meth}
We begin by providing an evaluation of several recent few-shot learning methods on LRDS. In particular, we focus on the best performing approaches from the recent study of Chen \textit{et al.}~\cite{closerlook}, and report the results of prototypical networks~\cite{protonet}, relational networks~\cite{yang2018learning}, and the linear and cosine classifier baselines that have shown promising results in~\cite{closerlook}. In addition, we evaluate the very recent Proto-MAML approach~\cite{triantafillou2020meta} which achieves top performance on the realistic MetaDataset. We train all the models under the full data setting and report 5-shot accuracy on the novel set in Table~\ref{table:comp-sota}.
\begin{table}
\centering
\resizebox{0.8\linewidth}{!}{
\begin{tabular}{l@{\hspace{8mm}}c@{\hspace{8mm}}c}
\hline
Model & Top-1 & Top-5  \\

\hline \hline
Linear Classifier~\cite{closerlook}   & 17.10 & 34.46 \\
Prototypical Networks~\cite{protonet} & 17.14 & 33.81 \\
Relational Networks~\cite{relation-net} & 8.34 & 25.53 \\
Cosine Classifier~\cite{closerlook}  & 13.65 & 32.33\\
Proto-MAML~\cite{triantafillou2020meta} & 14.54 & 30.37\\
\hline
\end{tabular}
}
\caption{Benchmarking few-shot learning methods. 5-shot accuracy for models trained with full data is reported.
}
\label{table:comp-sota}
\vspace{-15px}
\end{table}

\begin{table*}
\centering
\resizebox{0.80\linewidth}{!}{
\begin{tabular}{c@{\hspace{8mm}}l@{\hspace{8mm}}c@{\hspace{4mm}}c@{\hspace{4mm}}c@{\hspace{4mm}}c@{\hspace{4mm}}c} 
\hline
\multirow{3}{*}{\makecell[c]{Type of \\ supervision}} & \multirow{3}{*}{Model} & \multirow{3}{*}{Base-Val} & \multicolumn{4}{c}{Novel-test set}\\ 
& & &\multicolumn{2}{c}{1-shot} & \multicolumn{2}{c}{5-shot}  \\ 
\cline{4-7}
  &         &                    & Top-1 & Top-5   & Top-1   & Top-5     \\
\hline \hline
&Baseline            & 44.13     & 7.00 & 16.36  & 17.10  & 34.46    \\
\hline
Attributes &+Attribute            & 45.38     & 7.56 & 17.00  & 19.66  & 37.62    \\
\hline
\multirow{2}{*}{\makecell[c]{Class \\ Hierarchy}}&+Hierarchy Embedding  & 44.57     & 7.48 & 17.34  & 19.12  & 36.93    \\
&+Hierarchy Classifier   & 46.43     & 7.83 & 17.97  & 19.57  & 37.19    \\
\hline
Bounding Box &+Bounding Box         & 45.97     & 7.14 & 17.16  & 19.64  & 37.40    \\
\hline
\multirow{2}{*}{\makecell[c]{Segmentation}}&+Segmentation Region  & 45.68     & 7.68 & 17.35  & 18.95  & 37.46    \\
&+Segmentation FCN    & 45.82     & 7.84 & 17.53  & 20.02  & 38.26    \\
\hline
Scene Label & + Scene  & 45.03 & 7.37 & 18.08 & 18.26 & 36.45 \\
\hline
\end{tabular}}
\caption{Comparison of different supervision sources on the base-validation set and novel-test set of \benchmark. For clarity, the models are trained with full data.}
\label{table::sups}
\vspace{-15px}
\end{table*}
Similar to~\cite{closerlook}, we observe that prototypical networks show strong results; however, the gap between prototypical and relational networks is more significant on our benchmark. We hypothesize that this is due to the fact that relational networks do not generalize well to more realistic data distributions. A similar observation can be made for Proto-MAML. While this complex approach outperforms the linear classifier baseline on MetaDataset, it struggles LRDS, further illustrating the unique challenges presented by our benchmark. Based on these observations, we use linear classifier for the rest of the experiments in the paper.

\subsection{Exploration of Individual Supervision Sources}
\label{subsec::individual_sup}
We now explore a diverse set of supervision sources on the proposed LRDS benchmark, while using the full training set of the base classes. In Table~\ref{table::sups} we report results for 5 types of supervision, including attributes, class hierarchy, bounding boxes, segmentation, and scene labels. In  addition, we report results for a few  more sources of supervision, including self-supervised objectives like rotation and relative patch location in the appendix. 

\vspace{-12px}
\paragraph{Attributes.}
Following~\cite{tokmakov2018learning}, we use a multi-label classification loss for the attributes as shown in Figure~\ref{fig::model}. The results in the second row of Table~\ref{table::sups} demonstrate that this approach indeed improves the performance on the novel classes. In particular, top-5 classification performance in the 5-shot scenario is increased by more than 3\%,  and the improvements on the base set are a lot less significant, confirming the observations of ~\cite{tokmakov2018learning}.

\vspace{-12px}
\paragraph{Class Hierarchy.}
We first try to incorporate the hierarchical structure of categories into the feature space. Following~\cite{li2019large}, we utilize a hierarchical embedding space, transforming the feature representation to different levels of semantic hierarchy to classify corresponding concepts. 

We further experiment with a simplified version of class hierarchy supervision, that is, classifying an object on multiple concept levels. For instance, we can classify a cat as {\tt cat}, {\tt mammal}, and {\tt animal}. In implementation, we split the WordNet of ADE20K~\cite{ADE20K} into four levels and learn four independent classifiers on each of the levels. This naive approach outperforms the hierarchical embedding on LRDS, and we use it for the rest of the experiments.

\vspace{-12px}
\paragraph{Bounding Boxes.} Unlike~\cite{wertheimer2019few}, we study the effect of providing bounding box labels for the training on base classes. The intuition is that this will allow the object representation to focus on the objects and not on the background, thus generalizing better to unseen object distributions. To this end, we add a bounding box regression layer identical to R-CNN~\cite{girshick2014rich} after the RoI-Align stage. As can be seen from the corresponding row of Table~\ref{table::sups}, this results in a significant performance improvement on the novel categories, confirming our intuition.

\vspace{-12px}
\paragraph{Segmentation.} We experiment with two variants of providing segmentation supervision. The first version follows the protocol in Mask R-CNN~\cite{maskrcnn}, where we predict a binary mask inside an RoI-Aligned region. The second version operates in the semantic segmentation regime~\cite{long2015fully}, appending an additional convolutional layer after the feature map. Notice that we only apply segmentation supervision to the objects of base classes in the training set and ignore the other pixels. The numbers in Table~\ref{table::sups} demonstrate that both approaches result in an improvement over the baseline and the semantic segmentation model achieves top performance. We hypothesize that the problem of semantic labeling of the pixels in a large feature map is more difficult than the alternative formulation of binary pixel classification in a cropped region, providing a stronger regularization signal for representation learning.

\vspace{-12px}
\paragraph{Scene Labels.}
As shown in Figure~\ref{fig::model}, we supervise scene labels by applying a global pooling layer and linear scene classifier after the feature map of the image. Although scene labels are not directly related to the object categories, we observe a significant improvement in the novel classification performance. We attribute it to the fact that scenes are correlated with certain groups of object categories in the same way as class attributes or elements of the class hierarchy, regularizing learning of object representation.

Overall, the effect of additional supervision sources on the training process varies significantly with their types. We hypothesize that the difficulty of a supervisory task is directly related to how useful it is for learning generalizable representations. For instance, segmentation labels seem to be more useful than bounding boxes, as learning to predict the precise boundaries of the objects is harder than approximately localizing them with a box. The main implication of this observation is that investing in more expensive annotations does pay off when used for representation learning.

\subsection{Varying the Amount of Supervision}
\label{subsec::small_sup}

\begin{figure}
    \centering
    \includegraphics[width=1.0\columnwidth]{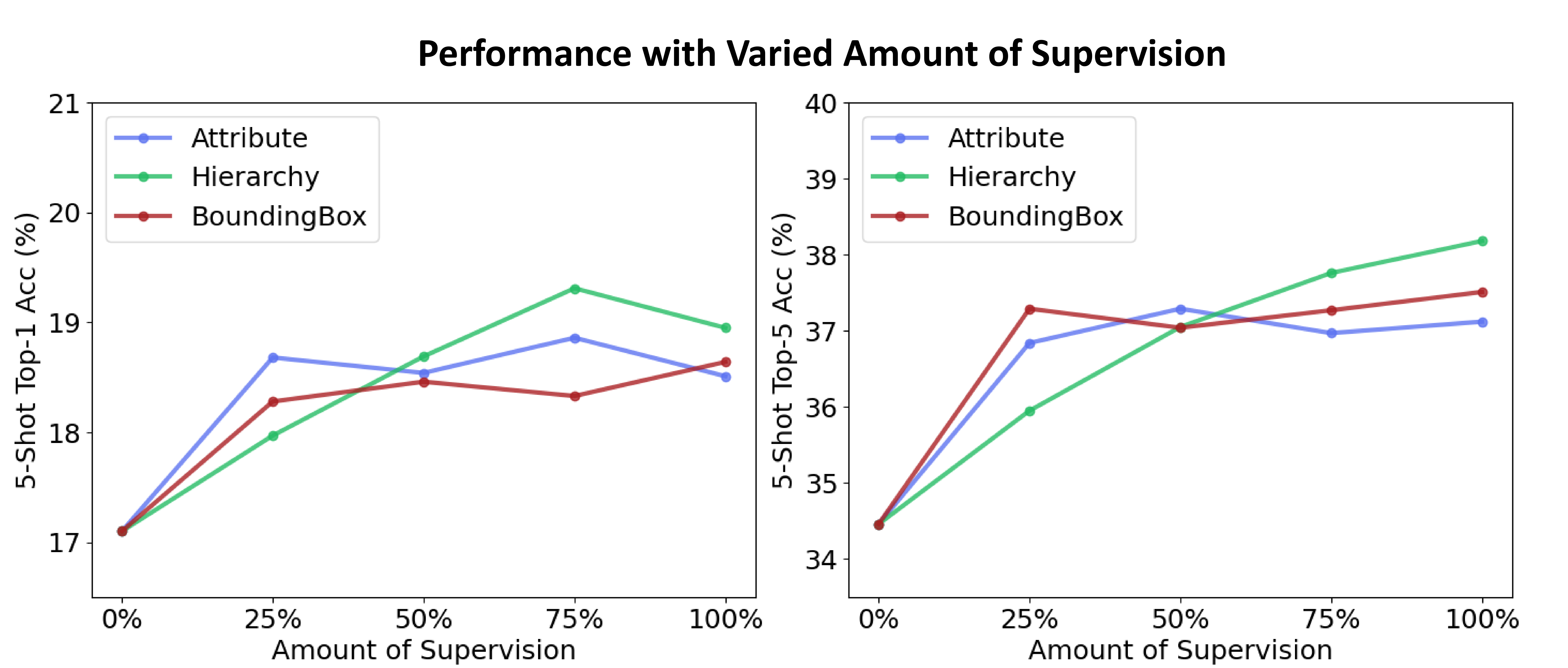}
    \caption{Varying the amount of supervision.  The  performance  increases when additional labels are available even for as little as 25\% of the instances.}
    \label{fig::supervision_ratio}
    \vspace{-15px}
\end{figure}

After observing that providing diverse sources of supervision can bring significant improvements in few-shot learning performance, we want to explore whether these improvements can \textit{be obtained at a lower cost}. In this section, we explore the effect of the fraction of data for which additional labels are provided on the quality of learned representation. In particular, we experiment with three kinds of supervision that have shown strong improvement: attributes, class hierarchy, and segmentation, varying the proportion of labeled data, and report the results in Figure~\ref{fig::supervision_ratio}.

We can observe that for all three types of supervision, labeling as little as 25\% of the data already results in significant improvements over the baseline. Indeed, labeling 25\% of instances already covers a lot of categories as well as different scenarios, providing a sufficiently strong regularization signal for representation learning. These results demonstrate that the benefits of diverse supervision can be obtained at a relatively low cost. For the rest of the paper, however, we use the full set of of annotations for each source of supervision to demonstrate the full potential of this approach.

\subsection{Combining Multiple Sources of Supervision} 
\label{subsec::mult_sup}
\begin{table}
	\centering
	\renewcommand{\arraystretch}{1.2}
	\renewcommand{\tabcolsep}{1.2mm}
	\resizebox{0.9\linewidth}{!}{
		\begin{tabular}{l@{\hspace{8mm}}c@{\hspace{8mm}}c@{\hspace{8mm}}c}
			\hline
			Model           & Method            & Top-1 & Top-5 \\\hline\hline
			Baseline        & \textbackslash{}  & 17.10 & 34.46 \\\hline
			+Seg            & MTL               & 18.43 & 35.49 \\ 
			+Seg            & CL                & 20.02 & 38.26 \\\hline
			+Seg+Attr       & MTL               & 19.59 & 37.99 \\ 
			+Seg+Attr       & CL                & 20.96 & 39.41 \\\hline
			+Seg+Hier       & CL                & 21.02 & 39.31 \\
			+Seg+BBox       & CL                & 20.98 & 39.14 \\ 
			+Seg+Hier+Attr  & CL                & 20.40 & 39.40 \\
			+Seg+Attr+Hier  & CL                & 21.18 & 39.99 \\ \hline
		\end{tabular}
	}
	\caption{Combining multiple types of supervision. We compare both methods for training on multiple tasks (MTL and CL), as well as different task combinations on the novel set of LRDS.}
	\label{table::mtl}
	\vspace{-15px}
\end{table}
It is natural to ask whether combining multiple sources of supervision can lead to further improvements. We answer this question in Table~\ref{table::mtl}, and begin by exploring two settings: multi-task learning (MTL), where all types of supervision are applied at once, and curriculum learning (CL), where the additional types of supervision are added one by one (the details of each setting are reported in the appendix). We observe that both for segmentation alone (combined with the main task of object classification), and for a combination of segmentation with attributes, CL achieves a better performance compared to MTL, demonstrating the complexity of multi-task learning. We thus use CL for combining multiple sources of supervision in the rest of the paper.

Next, we notice that combining diverse sources of supervision can indeed have an additive effect on the generalization abilities of the learned representation. For instance, adding both semantic and localization tasks on top of semantic segmentation noticeably improves the performance on the novel set. Moreover, combining three types of supervision results in further improvements, though the returns tend to diminish as more sources of supervision are added. 

Finally, the orders in which the sources of supervision are added in our CL framework influence the final performance. In Table~\ref{table::mtl}, two different ways of combining segmentation, attributes, and class hierarchy result in a noticeably different results on the novel set. Overall, we observe that adding stronger forms of supervision first leads to better results in our framework. We hypothesize that this is due to the fact that earlier stages of representation learning are more important, and later fine-tuning stages have only a minor influence. However, optimal rules for selecting this order remain to be thoroughly explored.

\subsection{Effect of Diverse Supervision on Small Data}
\label{sec::less_data}
\begin{figure}
    \centering
    \includegraphics[width=1.0\columnwidth]{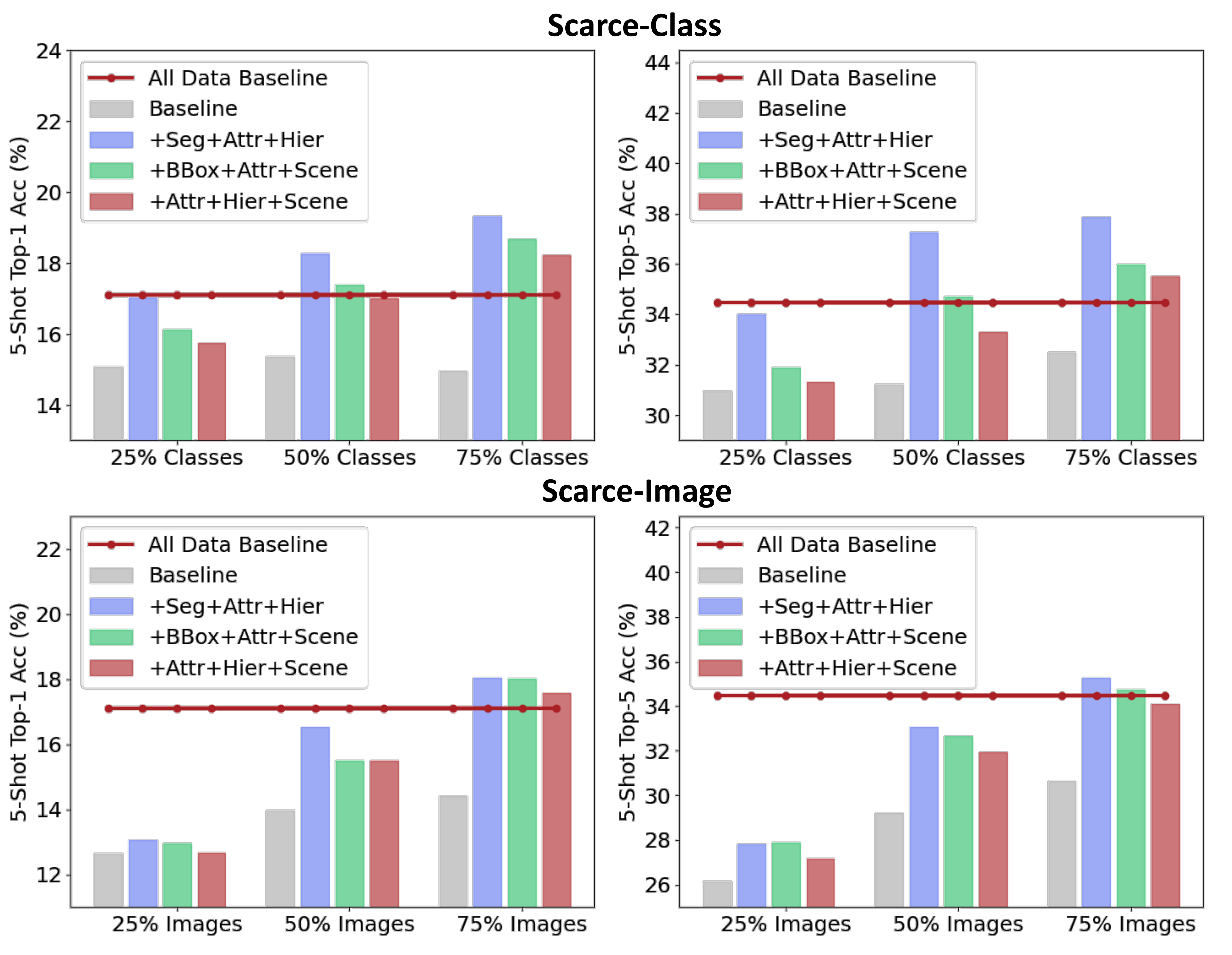}
    \caption{Effect of diverse supervision on small data. We follow the notation introduced in Figure~\ref{fig::cover}. Additional sources of supervision consistently improve the performance under all settings.}
    \label{fig::sup_enhance}
    \vspace{-15px}
\end{figure}

Finally, we answer the question raised at the beginning of this paper: ``Can diverse supervision remedy the small data constraints?'' To this end, we first establish the baselines for each setting (\FewC~and \FewI) and each data ratio (75\%, 50\%, and 25\%) by training on classification labels alone, and then augment those with different combinations of supervisory sources. The particular settings are chosen based on an extensive study of the optimal combinations in the appendix, and include both the most effective but expensive types of supervision (such as bounding box and segmentation) and more cost-effective labels (such as attributes and scene classes). The results are reported in Figure~\ref{fig::sup_enhance}.

First, we observe that decreasing both the number of base classes and the number of images results in a significant drop in the performance of the baseline, confirming the complexity of the small data setting studied in this paper. It is worth noting that the drop in performance is not uniform across the settings. Instead, the lack of images has a larger effect on the model's performance than the lack of classes. As discussed in Section~\ref{subsec::small_regime}, this is due to the fact that removing images reduces the actual number of training instances by a large margin, compared to removing the least frequent classes. We quantify this observation in Table~\ref{tab:inst_portion}. 

Next, we can see that various combinations of diverse supervision can indeed significantly improve the baseline performance in these challenging scenarios. For instance, in the \FewC~setting, our best combination of supervisory sources (Seg + Attr + Hier) outperforms the model trained on the full base training set by using only 50\% of the classes. As discussed above, the \FewI~setting is more challenging; however, diverse supervision still provides noticeable improvements, nearly reaching the the full data performance by only using half of the images. 

Overall, using both expensive and cheap sources of supervision results in improvements across the board. That said, we also observe that in the most challenging scenario, where only 25\% of the images are available, all the models struggle to learn a generalizable representation, emphasizing the complexity of the problem. In future work this limitation can be addressed by combining a small number of densely labeled images, with a large collection of unlabeled ones, using recent advances in self-supervised learning~\cite{he2020momentum}.



\begin{table}
    \centering
    \resizebox{0.7\linewidth}{!}{
    \begin{tabular}{cc}
        \hline
        Settings & Remaining Instance Portion \\
        \hline\hline
        25\% Images & 25.14\%\\
        50\% Images & 49.79\%\\
        75\% Images & 75.00\%\\
        \hline
        25\% Classes & 80.87\%\\
        50\% Classes & 92.73\%\\
        75\% Classes & 97.39\%\\
        \hline
    \end{tabular}}
    \caption{The portion of remaining samples under \FewC~and \FewI~settings. The former retains significantly fewer instances than the latter.}
    \label{tab:inst_portion}
    \vspace{-15px}
\end{table}

\vspace{-5px}
\section{Conclusion and Discussion}
\vspace{-5px}
In this paper we have introduced LRDS -- a realistic few-shot learning benchmark with rich annotations. In addition, we have proposed two new evaluation regimes -- \FewC\ and \FewI, which emulate the real-world issues of class and image scarcity, respectively. We have then explored how these challenging scenarios can be remedied by using diverse supervision available in LRDS. Our experiments have shown that a variety of supervision sources, as well as their combinations, are in fact helpful for the task. When combining multiple sources of supervision, those who are most different in nature, such as semantic and localization cues, are most complementary. However, we have also encountered some open problems which we discuss below.

First, the performance quickly saturates as more supervision sources are added. This is a common problem in multi-task learning, and a subject of an active research. We expect that as better approaches to training multi-task models are developed, they will automatically increase the value of diverse supervision for learning generalizable representations in the small data regime. 

Second, as is discussed in Section~\ref{subsec::mult_sup}, the order in which multiple sources of supervision are applied has a noticeable effect on performance. As the number of supervision sources grows, finding the optimal order in a naive way will become prohibitively expensive. Devising principled or heuristic rules to guide the search is thus very important. While we provide some preliminary intuition into this problem (\ie, ``harder'' tasks need to be learned first), it still remains to be thoroughly explored.

{\small
\bibliographystyle{ieee_fullname}
\bibliography{arxiv}
}

\newpage
\quad
\newpage
\appendix
\section*{Appendix}

This appendix provides additional experimental results and details. We summarize and highlight the relationship and difference with related work in Section~\ref{sup::sec::related}. We include additional implementation and experimental details in Section~\ref{sup::sec::details}. We provide additional experimental results and analysis about the ``Small Data'' scenarios in Section~\ref{sup::sec::small_data}, including combining multiple sources of supervision and comparing the \FewC~and \FewI~settings. We include extensive evaluation on all the supervision sources in Section~\ref{sup::sec::supervision}. We show the cross-domain generalization of the learned representation on other datasets in Section~\ref{sup::sec::cross}. We demonstrate the consistent usefulness of incorporating diverse supervision into other few-shot learning methods in Section~\ref{sup::sec::few_shot}. We include the results of pre-trained models based on self-supervision loss in Section~\ref{sec::pretrain}. We provide the baseline of using object crops for classification in Section~\ref{sec::obj_crop}. We present experiments using advanced multi-task learning methods in Section~\ref{sec::advance_mtl}. Finally for completeness, we provide the corresponding numbers for the experimental results shown in the form of figures in Section~\ref{sup::sec::quant}.

\section{Summary of Connection and Difference with Related Work}
\label{sup::sec::related}
Our work is broadly related to the general investigation of learning with varying amount of data and annotation. While the detailed discussion on our proposed settings of \FewC~and \FewI~and existing work has been already covered in Section~\main{2} of the main paper, here we {\em further summarize and highlight} their connection and difference in Table~\ref{sup::tab::diff}. As shown in Table~\ref{sup::tab::diff}, our work is unique and addresses significant limitations in existing work.

\begin{table*}[htbp]
    \centering
    \resizebox{1.0\linewidth}{!}{
    \begin{tabular}{c|c|ccc|cc}
        \hline
        \multirow{3}{*}{Related Settings} & \multirow{3}{*}{Included Tasks} & \multicolumn{3}{c}{Data} & \multicolumn{2}{c}{Annotations}\\
        \cline{3-7}
        & & \makecell{Amount for\\Feature Learning} & \makecell{Data Distribution for \\ Feature Learning} & Test on Novel & Amount & Types \\
        \hline\hline
        \makecell{\FewC~and\\ \FewI} & \makecell{Single target task \\ Multiple supervisory tasks} & Small & Imbalanced & \cmark & Full & Multiple \\\hline
        Few-shot Learning & Single target task & Large & \textbackslash{} & \cmark & Full & Single \\\hline
        Multi-task Learning & Multiple target tasks & Large & \textbackslash{} & \xmark & Full & Multiple \\\hline
        Long-tail Learning & Single target task & Large & Imbalanced & \xmark & Full & Single\\\hline
        Weakly-supervised Learning & Single target task & Large & \textbackslash{} & \xmark & Partial & Single\\\hline
        Unsupervised Learning & Single target task & Large & \textbackslash{} & \xmark & None & None \\\hline
    \end{tabular}}
    \vspace{.3em}
    \caption{Summary of the commonalities and differences between our proposed settings of \FewC~and \FewI~in \benchmark~and existing work on learning with varying amounts of data  and annotation. `\textbackslash{}' means that the setting poses no requirements on whether the training data should be balanced or imbalanced.}
    \label{sup::tab::diff}
\end{table*}

\section{Additional Implementation Details}
\label{sup::sec::details}

\subsection{Benchmark Construction}
In this section, we include additional details for constructing our \benchmark~benchmark, especially the algorithms for box enlargement and random jittering (Algorithm~\ref{sup::algo::enlarge}), which were briefly discussed in Section~\main{3.2} of the main paper.

First, we enlarge the bounding box size by a context ratio $\gamma$, which is 2.7, computed from the average ratio between the tight bounding boxes and full images on ImageNet~\cite{imagenet}. Then in function $\mathtt{RatioAssign}$, we assign the enlargement in height $h$ and width $w$ by $\gamma_h$ and $\gamma_w$, respectively. Note that $\gamma_h$ and $\gamma_w$ are generated randomly. They are larger than 1 and $\gamma_h \gamma_w = \gamma$.

Then in function $\mathtt{FindJitterRange}$, we compute the range of movement for jittering the bounding box center, including the maximum range on the y-axis and x-axis $y_{min}, y_{max}, x_{min}, x_{max}$. The jittered box should still be inside the image and contain the original tight bounding box.

Finally, we randomly sample the jittering movement $m_x, m_y$ from the computed range and apply it to the bounding box center, thus finishing the box enlargement and jittering.
\vspace{-10px}
\begin{algorithm}
    \renewcommand{\algorithmicrequire}{\textbf{Input:}}
	\renewcommand{\algorithmicensure}{\textbf{Output:}}
    \caption{Enlarging and Jittering Boxes}
    \label{algo::enlarge}
    \begin{algorithmic}[1]
    \REQUIRE Image Size: $(H ,W)$, Tight BBox Size: $(h, w)$, BBox Center $(y, x)$, Context Ratio $\gamma$
    \ENSURE BBox for \benchmark, including: Size $(\widetilde{h}, \widetilde{w})$,\\ Center $(\widetilde{y}, \widetilde{x})$
    \STATE $\gamma_h, \gamma_w \xleftarrow{} \mathtt{RatioAssign}(H, W, h, w, y, x, \gamma)$
    \STATE $\widetilde{h} \xleftarrow{} h\gamma_h$, $\widetilde{w} \xleftarrow{} w\gamma_w$
    \STATE $y_{min}, y_{max}, x_{min}, x_{max}\xleftarrow{} \mathtt{FindJitterRange}(H, W, h, w, \widetilde{h}, \widetilde{w}, y, x) $
    \STATE $m_x \xleftarrow{} \mathtt{UniformSampling}(x_{min}, x_{max})$
    \STATE $m_y \xleftarrow{} \mathtt{UniformSampling}(y_{min}, y_{max})$
    \STATE $\widetilde{y}\xleftarrow{} y + m_y, \widetilde{x}\xleftarrow{} x + m_x$
    \STATE Return $\widetilde{h}, \widetilde{w}, \widetilde{y}, \widetilde{x}$
    \end{algorithmic}
    \label{sup::algo::enlarge}
\end{algorithm}
\vspace{-10px}

\subsection{Model Training}
\label{sup::subsec::train}
In this section, we provide the implementation details for the experiments in Section~\main{5} of the main paper.
Under the backbone of ResNet-18~\cite{resnet}, we experimented with varied implementations of the model and different hyper-parameter settings. In the process of data loading, we resize all the short edges of the images to the length of 800 following the protocol in the ADE20K dataset~\cite{ADE20K}. As for the model, we modify the down-sampling rate of ResNet-18, with the first three Residual Blocks yielding a down-sampling rate of 2, which together down-samples the image by 8, compared to 32 in original ResNet. During the training of the model, we use a batch size of 8, optimizer of SGD with learning rate 0.1, and cosine scheduler~\cite{loshchilov2016sgdr}. The whole training process of the baseline model takes 6 epochs to run, roughly 3 hours on a 4-GPU machine.

As for the combination weights for different types of supervision, we adjust the weights so that the total loss of each supervision branch has the same scale as the classification branch. The detailed values of the weight hyper-parameters are summarized in Table~\ref{sup::tab::weight}.

\begin{table}[htbp]
    \centering
    \resizebox{0.5\linewidth}{!}{
    \begin{tabular}{l@{\hspace{5mm}}c@{\hspace{5mm}}}
    \hline
        Supervision Type & Weight \\\hline\hline
        Attribute & 25.0\\
        Hierarchy & 1.0\\
        Scene & 0.2\\
        Part & 25.0\\
        Bounding Box & 5.0\\
        Segmentation & 0.5\\
        Rotation & 10.0\\
        Patch Location & 1.0\\\hline
    \end{tabular}}
    \vspace{.3em}
    \caption{Weights for different types of supervision during training.}
    \label{sup::tab::weight}
\end{table}

\subsection{Training Few-shot Learning Methods}

In this section, we provide the details for experimenting the few-shot learning methods in Section~\main{5.1} of the main paper. During the feature representation learning stage, we train the models following Section~\ref{sup::subsec::train}. Then during the stage of few-shot learning on novel classes, we append and train an additional linear layer on top of the learned features.

For Cosine Classifier~\cite{closerlook}, we simply replace the linear classifier in our baseline model with a cosine classifier. For Prototypical Network~\cite{protonet}, we use a full 193-way, 5-shot (or 1-shot) support set to calculate the mean of each category and then perform 193-way classification on a query set. We reproduce Relational Network~\cite{relation-net} in a setting similar to Prototypical Network. For Proto-MAML~\cite{triantafillou2020meta}, we first add one linear layer after the fixed feature extractor as an additional encoder. We use the 100-way novel-val set to estimate initialization parameters for the encoder. Then we evaluate model performance on the 193-way novel-test set. On both novel-val set and novel-test set, we use a prototypical network initialization as is described in~\cite{triantafillou2020meta}.

\section{Additional Analysis on ``Small Data'' Scenarios}
\label{sup::sec::small_data}
This section provides additional experimental evaluation and analysis in the ``Small Data'' regimes.

\subsection{Combining Multiple Sources of Supervision}
\begin{figure}
	\includegraphics[width=1.0\columnwidth]{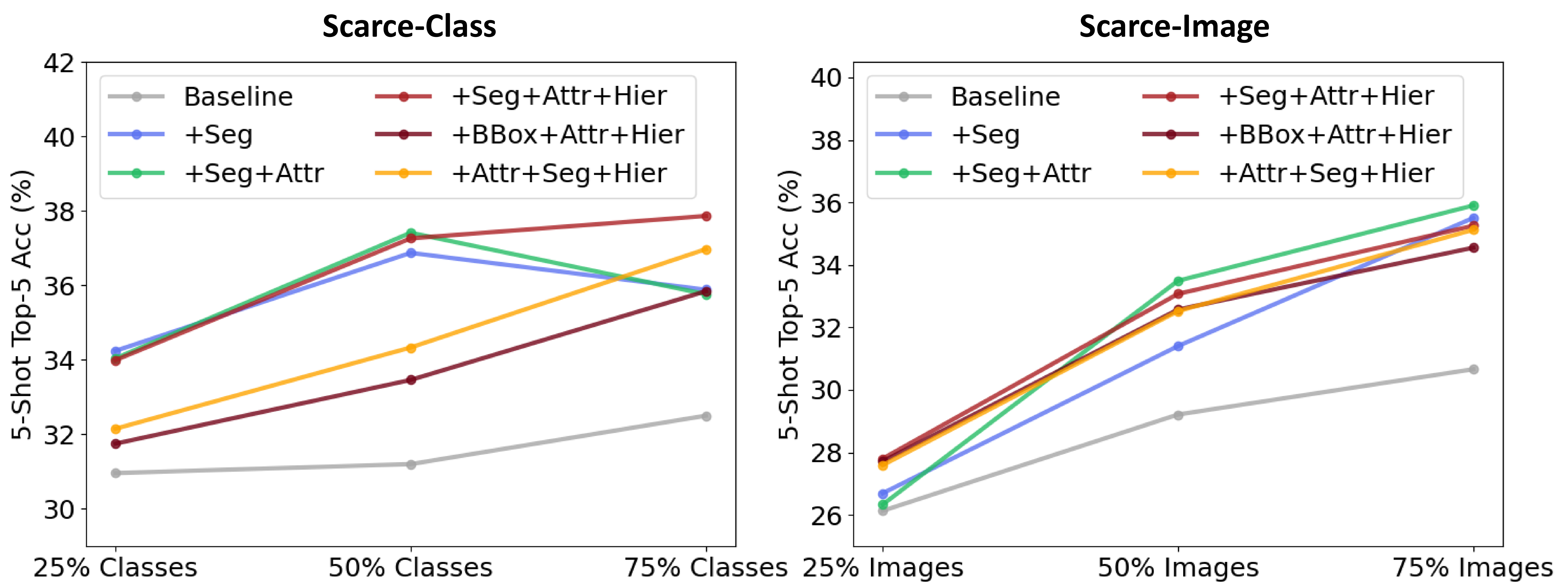}
	\caption{Combining multiple sources of supervision consistently improves the performance under \FewC~and \FewI~settings.}
	\label{sup::fig::multi_sup}
\end{figure}

In Section~\main{5.4} of the main paper, we showed that combining multiple sources of supervision leads to improvements in full data settings. In this section, we further demonstrate the {\em consistent effectiveness} of diverse supervision in the very challenging ``Small Data'' regimes. As summarized in Figure~\ref{sup::fig::multi_sup}, incorporating supervision improves the performance of the baseline model by large margins, even with very few classes or images.

In addition, we have the following observations from Figure~\ref{sup::fig::multi_sup}. First, we observe a significant performance drop when we replace the segmentation supervision with bounding box supervision. This is due to the fact that segmentation provides a stronger learning signal by exactly delineating the objects from the background, in contrast to approximately localizing them with bounding boxes. This further confirms that investing in more expensive types of annotations is valuable, especially when the number of training instances is limited.

Second, we can see that when the number of training samples is decreased, the improvements from additional supervision sources decrease. This demonstrates the limitations of not only our approach, but also modern representation learning techniques in general, since they struggle in extremely low data regimes. That said, using additional sources of supervision leads to significant improvements at moderate data scarcity levels.

Finally, we find that the performance of the ``+Seg'' and ``+Seg+Attr'' models drops in the Scarce-Class evaluation  when the number of available classes increases from 50\% to 75\%. A potential explanation for this lies in the long-tail distribution of \benchmark. Removing the tail classes from the training set makes the category-level instance distribution more balanced, thus simplifying the optimization. 

\subsection{Comparison Between \FewI~and \FewC~Settings}
\begin{figure}
	\includegraphics[width=1.0\columnwidth]{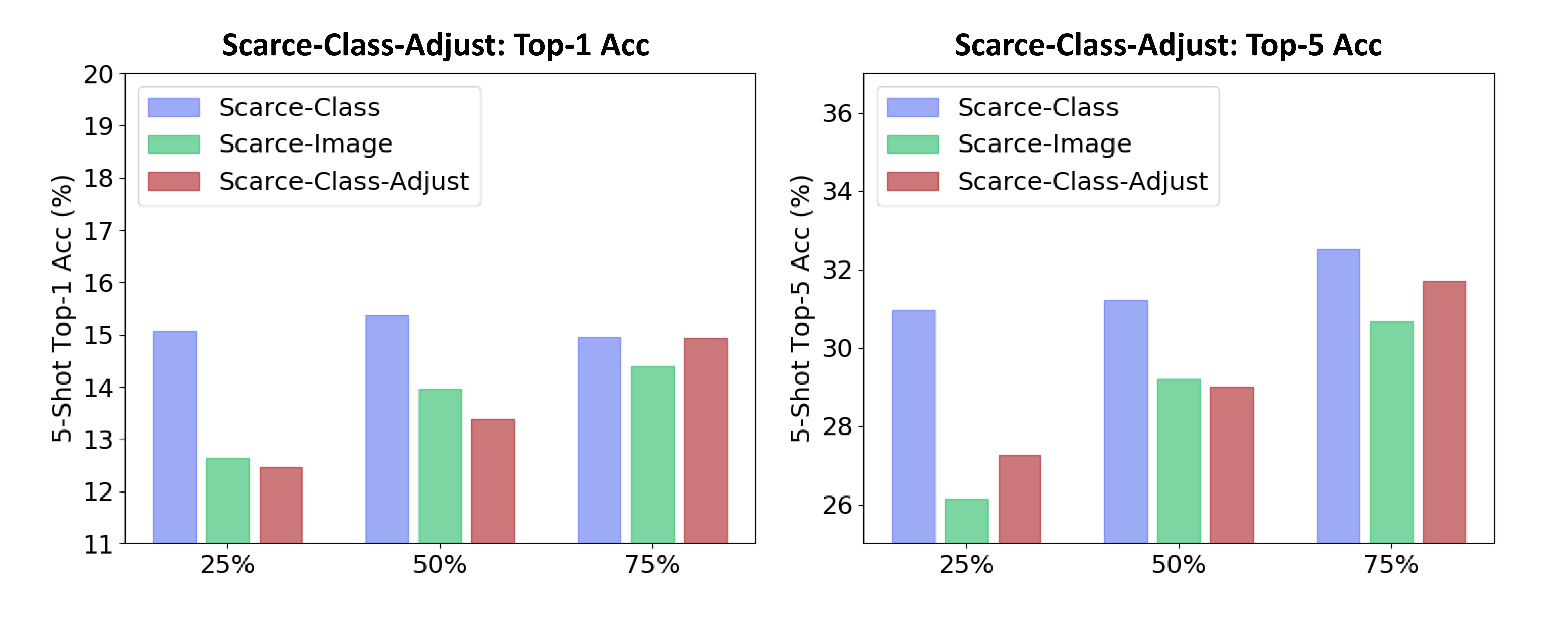}
	\caption{Performance comparison of the baseline model under \FewC, Scarce-Class-Adjust, and \FewI~settings. The new setting ``Scarce-Class-Adjust'' is constructed through randomly down-sampling the instances of each category in the original \FewC~setting, so that the total number of training samples is the same as that of the \FewI~setting. With the same amount of training data, the lack of classes has a similar effect as the lack of images on the performance drop of the baseline.
	}
	\label{sup::fig::adjust}
\end{figure}

In Section~\main{5.5} and Table~\main{4} of the main paper, we showed that the lack of images (\FewI) has a larger effect than the lack of classes (\FewC) on the performance drop of the baseline. This is attributed to the fact that removing images reduces the actual number of training instances by a large margin, compared to removing the least frequent classes. To further illustrate this observation, we construct a new setting ``Scarce-Class-Adjust,'' through randomly down-sampling the instances of each category in the original \FewC~setting, so that the total number of training samples is the same as that of the \FewI~setting. As shown in Figure~\ref{sup::fig::adjust}, the baseline's performance is comparable in the settings of Scarce-Class-Adjust and \FewI.

\section{Additional Exploration of Individual Supervision Sources}
\label{sup::sec::supervision}
We demonstrated the effectiveness of five representative types of supervision in Section~\main{5.2} of the main paper. In this section, we provide {\em extensive} evaluation on all the supervision sources, including semantic supervision in Section~\ref{sup::sec::sem_sup}, localization supervision in Section~\ref{sup::sec::loc_sup}, and self-supervision in Section~\ref{sup::sec::self_sup}.

\begin{table*}[htbp]
\centering
\resizebox{0.9\linewidth}{!}{
\begin{tabular}{cc@{\hspace{8mm}}l@{\hspace{8mm}}c@{\hspace{4mm}}c@{\hspace{4mm}}c@{\hspace{4mm}}c@{\hspace{4mm}}c} 
\hline
\multirow{3}{*}{\makecell{Row\\Number}} & \multirow{3}{*}{Type of supervision} & \multirow{3}{*}{Model} & \multirow{3}{*}{Base-val} & \multicolumn{4}{c}{Novel-test set}\\ 
& & & &\multicolumn{2}{c}{1-shot} & \multicolumn{2}{c}{5-shot}  \\ 
\cline{5-8}
  &         &                &    & Top-1 & Top-5   & Top-1   & Top-5     \\
\hline \hline
1 & & Baseline               & 44.13 & 7.00 & 16.36 & 17.10 & 34.46    \\
\hline
2 & \multirow{5}{*}{Semantic supervision}
& +Attribute            & 45.38 & 7.56 & 17.00 & 19.66 & 37.62\\
3 & & +Hierarchy Embedding  & 44.57 & 7.48 & 17.34 & 19.12 & 36.93\\
4 & & +Hierarchy Classifier & 46.43 & 7.83 & 17.97 & 19.57 & 37.19\\
5 & & +Scene                & 45.03 & 7.37 & 18.08 & 18.26 & 36.45\\
6 & & +Part                 & 45.68 & 7.39 & 17.23 & 19.39 & 37.2\\
\hline

7 & \multirow{5}{*}{Localization supervision}
& +Bounding Box         & 45.97 & 7.14 & 17.16 & 19.64 & 37.40\\
8 & & +Segmentation Region  & 45.68 & 7.68 & 17.35 & 18.95 & 37.46\\
9 & & +Segmentation FCN     & 45.82 & 7.84 & 17.53 & 20.02 & 38.26\\
10 & & +Stuff                & 43.86 & 6.58 & 15.04 & 15.63 & 32.65\\
11 & & +(Object+Background)  & 46.03 & 7.09 & 16.92 & 20.37 & 38.55\\
\hline
12 & \multirow{2}{*}{Self-supervision}
&+Rotation              & 44.31 & 6.16 & 15.16 & 17.57 & 35.04\\
13 & &+Patch Location        & 44.43	& 7.28 & 16.64 & 18.49 & 35.53\\
\hline
\end{tabular}}
\vspace{.3em}
\caption{Comparison of different supervision sources on the base-validation set and novel-test set of \benchmark. The models are trained with full data. All types of supervision are effective by themselves (except stuff supervision which needs to be combined together with foreground supervision). And annotated semantic and localization supervision outperforms self-supervision.}
\label{sup::tab::sups}

\end{table*}

\subsection{Semantic Supervision}
\label{sup::sec::sem_sup}

In Section~\main{5.2} of the main paper, we discussed the types of supervision that leverage semantic information: ``Attributes,'' ``Class Hierarchy,'' and ``Scene Labels.'' Here we further investigate another type of semantic supervision: ``Object Parts.''

\paragraph{Object parts.}
Similar to the attributes, we use a multi-label classification loss for the object parts as shown in Figure~\main{4} of the main paper. The ground-truth of the object parts comes from the part annotation in ADE20K~\cite{ADE20K}. The results in row 6 of Table~\ref{sup::tab::sups} indicate that part labels also result in improved generalization performance, though the improvement is not as significant as that of the other types of semantic supervision.

\subsection{Localization Supervision}
\label{sup::sec::loc_sup}
In Section~\main{5.2} of the main paper, we discussed the types of supervision that leverage location information: ``Segmentation'' and ``Bounding Boxes.'' Here we further investigate how to use the background segmentation information, namely ``Stuff Segmentation.''

\paragraph{Stuff segmentation.}
For stuff segmentation, we follow FCN~\cite{long2015fully} and append a convolutional layer after the feature map, predicting a binary label for whether a pixel is object or stuff. Note that the pixels not belonging to the training set are still marked as unknown. The results in row 10 of Table~\ref{sup::tab::sups} show a small decrease in performance with respect to the baseline. We hypothesize that this is because the stuff supervision forces the representation to focus on the background features, which the object classifier then latches onto. 

We further combine the foreground and stuff labels together, giving a weight of 0.1 to the background classes. This combined supervision results in a performance improvement (row 11 in Table~\ref{sup::tab::sups}), outperforming even the variant with foreground segmentation only. This result demonstrates that stuff supervision can still be helpful, but only when combined with foreground supervision.

\subsection{Self-supervision}
\label{sup::sec::self_sup}
Figure~\main{4} in the main paper demonstrated that self-supervision can be naturally incorporated in our framework. In this section, we discuss in detail the effect of two widely used types of self-supervision, including ``Rotation'' and ``Relative Patch Location.''

\paragraph{Rotation.} We follow the pretext task in~\cite{gidaris2018unsupervised} by rotating the input image $\left \{ 0^{\circ},90^{\circ},180^{\circ},270^{\circ}\right \}$, and training an additional classifier to predict the angle of rotation. This method leads to improvement on the learned representation (row 12 of Table~\ref{sup::tab::sups}), but it is much smaller than that comes from annotated supervision.

\paragraph{Relative patch location.} Following~\cite{doersch2015unsupervised}, we divide the input image into a $3\times 3$ grid of crops. Then the center crop and another randomly picked crop are passed though a model to predict their relative locations. This self-supervision also effectively regularizes representation learning, as shown in row 13 of Table~\ref{sup::tab::sups}. Similar to the rotation supervision, the improvement still cannot match the other types of annotated supervision.

\section{Cross-Domain Generalization}
\label{sup::sec::cross}

In addition to the main experiments on \benchmark, we further investigate the generalization of the learned representation on ImageNet~\cite{imagenet}. Specifically, we use the feature representations trained on the base set of \benchmark~with and without additional supervision sources, and learn a linear classifier on top of them for the few-shot split of ImageNet defined in~\cite{hariharan2017low}. From the results in Table~\ref{table::transfer}, we observe that using additional sources of supervision also leads to improvements in a different domain, indicating a more generalizable representation.

\begin{table}[]
\centering
\resizebox{0.80\linewidth}{!}{
\begin{tabular}{l@{\hspace{8mm}}c@{\hspace{8mm}}c}
\hline
Model &  Top-1 & Top-5             \\ \hline\hline
Baseline        & 7.22  & 19.79 \\
+Attribute      & 8.06 & 21.40 \\ 
+Bounding Box   & 7.49  & 21.04 \\ 
+Class Hierarchy & 7.59 & 20.78 \\
+Segmentation   & 8.27 &  21.96     \\ 
+Seg+Attr & 8.49 & 22.86      \\ 
+Seg+Attr+Hie & 8.96 & 23.17 \\
\hline
\end{tabular}
}
\vspace{.3em}
\caption{Investigation of cross-domain generalization of the feature representation trained on \benchmark~for few-shot classification on ImageNet. Leveraging additional sources of supervision leads to more generalizable representation and thus improves the performance on ImageNet as well.}
\label{table::transfer}
\end{table}

\section{Effect on Other Few-shot Learning Methods}
\label{sup::sec::few_shot}
\begin{figure}
    \centering
    \includegraphics[width=\linewidth]{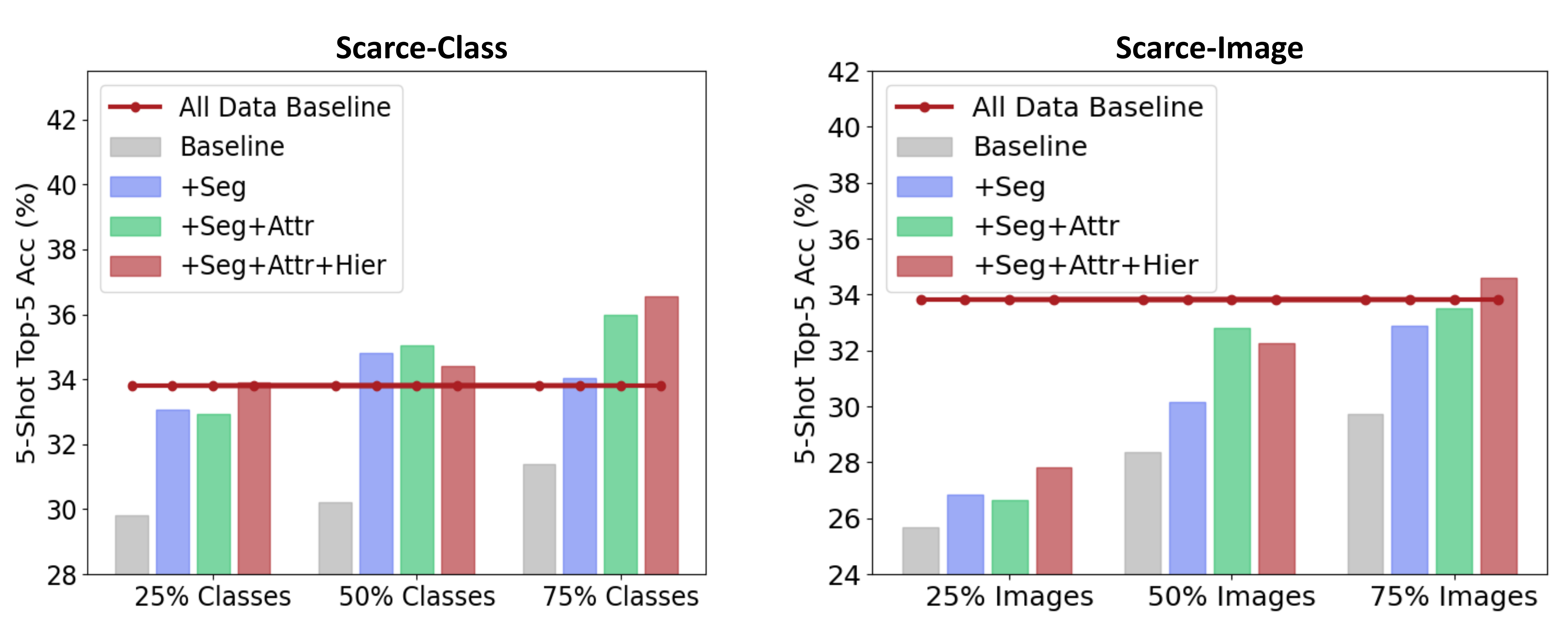}
    \caption{Incorporating additional supervision is consistently effective, which also improves the performance of the Prototypical Network baseline.
    }
    \label{sup::fig::proto}
\end{figure}

We mainly focused on adding supervision on top of the linear classifier baseline in Section~\main{5}. Here in Figure~\ref{sup::fig::proto}, we show the {\em consistent usefulness} of incorporating diverse supervision into other few-shot learning methods, such as Prototypical Network~\cite{protonet}.

\section{Pre-training with Self-supervision}
\label{sec::pretrain}
\textcolor{black}{
In this section, we provide the performance for the models pre-trained with a self-supervised pretext task: rotation loss~\cite{gidaris2018unsupervised}. We first train the model with the rotation branch only, and then add the class labels for fine-tuning. To study the effect of additional supervision on this model, we further incorporate several supervision sources to the model. As demonstrated in Table~\ref{table::pretrain}, the model pre-trained with the rotation self-supervision performs better than the baseline of training with class labels only, which is the result of better initialization. However, when adding more supervision ``Seg'' and ``Attr,'' while the performance of this variant still improves, the accuracy is lower than that of the model initialized with classification pre-training. This is a counter-intuitive observation, which demonstrates that self-supervised objectives are not always beneficial for representation learning. }

\begin{table}[]
\centering
\resizebox{0.70\linewidth}{!}{
\begin{tabular}{l@{\hspace{8mm}}c@{\hspace{8mm}}c}
\hline
Model &  Top-1 & Top-5             \\ \hline\hline
PT-Baseline & 17.41	& 36.05 \\
+Segmentation & 17.59 &	36.64\\
+Seg+Attr & 17.82	& 35.93      \\\hline
Baseline & 17.10 & 34.46\\
+Segmentation & 20.02 & 38.26 \\
+Seg+Attr & 20.96 & 39.41 \\
\hline
\end{tabular}
}
\vspace{.3em}
\caption{Effect of self-supervision pre-training. PT-baseline denotes the model initialized with a set of self-supervision pre-trained parameters. Baseline denotes the randomly initialized model.}
\label{table::pretrain}
\end{table}

\section{Baseline of Object Crops}
\label{sec::obj_crop}
\textcolor{black}{
In addition to the Faster R-CNN~\cite{FasterRCNN} based architecture in the main paper, we experiment with the most basic form of object classification, where the models operate on individual object crops instead of the whole feature map. Specifically, we crop each object out, resize each individual crop to $224 \times 224$, and then apply a CNN with global average pooling on it. 
From the numbers in Table~\ref{table::obj_crop}, this baseline achieves better performance than the Faster R-CNN~\cite{FasterRCNN} based architecture. This is because many objects are in small scales, and the operations of cropping and resizing enlarge them greatly.
}
\begin{table}[]
\centering
\resizebox{0.65\linewidth}{!}{
\begin{tabular}{l@{\hspace{8mm}}c@{\hspace{8mm}}c}
\hline
Model &  Top-1 & Top-5             \\ \hline\hline
Baseline & 21.13 & 42.51 \\
+Attr & 22.64 & 44.25\\
+Attr+Hie & 23.43 & 45.34\\
\hline
\end{tabular}
}
\vspace{.3em}
\caption{Object crop baseline model. Since the input object size is fixed, the performance is higher than the Faster R-CNN based model.}
\label{table::obj_crop}
\end{table}

\section{Advanced Multi-task Learning Methods}
\label{sec::advance_mtl}
\textcolor{black}{
To explore more ways of combining multiple sources of supervision, we experiment with a more complicated multi-task learning (MTL) method from ~\cite{kendall2018multi}. However, adding this new term does not improve the final classification performance, as shown in Table~\ref{table::advance_mtl}. This is due to the difference between the objectives of our problem and MTL: we are interested in improving the classification performance on novel categories, whereas MTL is concerned with the joint improvement on all the tasks.
}
\begin{table}[]
\centering
\resizebox{0.8\linewidth}{!}{
\begin{tabular}{l@{\hspace{8mm}}c@{\hspace{8mm}}c}
\hline
Model &  Top-1 & Top-5             \\ \hline\hline
Baseline & 17.10 & 34.46\\
+Seg+Attr~\cite{kendall2018multi} & 18.56 & 37.02\\
+Seg+Attr+Hie~\cite{kendall2018multi} & 18.16 & 36.92\\
\hline
+Seg+Attr (Ours) & 20.96 & 39.41\\
+Seg+Attr+Hie (Ours) & 21.18 & 39.99\\
\hline
\end{tabular}
}
\vspace{.3em}
\caption{Applying advanced MTL method in our model. Since we evaluate our model only by classification performance, using advanced MTL method does not lead to further improvement.}
\label{table::advance_mtl}
\end{table}

\section{Quantitative Results for Experiments}
\label{sup::sec::quant}
For those experimental results shown in the form of figures in the main paper and this appendix, here for completeness we provide their corresponding numbers in Tables~\ref{sup::tab::vary},~\ref{sup::tab::diverse_scarce_class},~\ref{sup::tab::diverse_scarce_image},~\ref{sup::tab::fig1left},~\ref{sup::tab::fig1right},~\ref{sup::tab::fig2},~\ref{sup::tab::fig4l}, and~\ref{sup::tab::fig4r}.

\label{sup::sec::vary}
\begin{table}[htbp]
    \centering
    \resizebox{0.85\linewidth}{!}{
    \begin{tabular}{cc|cc}
    \hline
        Type of supervision & Portion & Top-1 & Top-5 \\
        \hline\hline
        None(Baseline) & \textbackslash{} & 17.1 & 34.46\\\hline
        Attribute & 25\% & 18.68 & 36.84\\
        Attribute & 50\% & 18.64 & 37.29\\
        Attribute & 75\% & 18.86 & 36.97\\
        Attribute & 100\% & 18.51 & 37.12\\\hline
        Hierarchy & 25\% & 17.97 & 35.95\\
        Hierarchy & 50\% & 18.69 & 37.05\\
        Hierarchy & 75\% & 19.31 & 37.76\\
        Hierarchy & 100\% & 18.95 & 38.18\\\hline
        BoundingBox & 25\% & 18.64 & 37.51\\
        BoundingBox & 50\% & 18.33 & 37.27\\
        BoundingBox & 75\% & 18.46 & 37.04\\
        BoundingBox & 100\% & 18.26 & 37.29\\\hline
    \end{tabular}}
    \vspace{.3em}
    \caption{Performance of varied amount of supervision. Table for Figure~\main{5} in the main paper.}
    \label{sup::tab::vary}
\end{table}

\begin{table}[htbp]
    \centering
    \resizebox{1.0\linewidth}{!}{
    \begin{tabular}{cc|cc}
    \hline
        Model & Portion of Classes & Top-1 & Top-5 \\
        \hline\hline
        Baseline-Full Classes & 100\% & 17.10 & 34.46\\\hline
        Baseline & 25\% & 15.08 & 30.96\\
        Baseline & 50\% & 15.36 & 31.20\\
        Baseline & 75\% & 14.96 & 32.50\\\hline
        +Seg+Attr+Hier & 25\% & 17.02 & 33.99\\
        +Seg+Attr+Hier & 50\% & 18.26 & 37.26\\
        +Seg+Attr+Hier & 75\% & 19.30 & 37.86\\\hline
        +BBox+Attr+Scene & 25\% & 16.11 & 31.90\\
        +BBox+Attr+Scene & 50\% & 17.37 & 34.71\\
        +BBox+Attr+Scene & 75\% & 18.67 & 35.98\\\hline
        +Attr+Hier+Scene & 25\% & 15.73 & 31.32\\
        +Attr+Hier+Scene & 50\% & 16.99 & 33.31\\
        +Attr+Hier+Scene & 75\% & 18.21 & 35.52\\\hline
    \end{tabular}}
    \vspace{.3em}
    \caption{The effect of supervision combinations under the \FewC~setting. Table for Figure~\main{6} in the main paper, top row.}
    \label{sup::tab::diverse_scarce_class}
\end{table}

\begin{table}[htbp]
    \centering
    \resizebox{0.9\linewidth}{!}{
    \begin{tabular}{cc|cc}
    \hline
        Model & Portion of Images & Top-1 & Top-5 \\
        \hline\hline
        Baseline-Full Images & 100\% & 17.10 & 34.46\\\hline
        Baseline & 25\% & 12.64 & 26.13\\
        Baseline & 50\% & 13.96 & 29.21\\
        Baseline & 75\% & 14.39 & 30.66\\\hline
        +Seg+Attr+Hier & 25\% & 13.05 & 27.80\\
        +Seg+Attr+Hier & 50\% & 16.52 & 33.07\\
        +Seg+Attr+Hier & 75\% & 18.03 & 35.25\\\hline
        +BBox+Attr+Scene & 25\% & 12.95 & 27.88\\
        +BBox+Attr+Scene & 50\% & 16.50 & 32.66\\
        +BBox+Attr+Scene & 75\% & 18.01 & 34.75\\\hline
        +Attr+Hier+Scene & 25\% & 12.65 & 27.16\\
        +Attr+Hier+Scene & 50\% & 15.50 & 31.93\\
        +Attr+Hier+Scene & 75\% & 17.56 & 34.08\\\hline
    \end{tabular}}
    \vspace{.3em}
    \caption{The effect of supervision combinations under the \FewI~setting. Table for Figure~\main{6} in the main paper, bottom row.}
    \label{sup::tab::diverse_scarce_image}
\end{table}

\begin{table}[htbp]
    \centering
    \resizebox{0.9\linewidth}{!}{
    \begin{tabular}{cc|cc}
    \hline
        Model & Portion of Classes & Top-1 & Top-5 \\
        \hline\hline
        Baseline & 25\% & 15.08 & 30.96\\
        Baseline & 50\% & 15.36 & 31.20\\
        Baseline & 75\% & 14.96 & 32.50\\\hline
        +Seg    & 25\% & 17.31 & 34.24\\
        +Seg    & 50\% & 18.76 & 36.87\\
        +Seg    & 75\% & 18.11 & 35.88\\\hline
        +Seg+Attr   & 25\% & 17.27 & 34.04\\
        +Seg+Attr   & 50\% & 19.38 & 37.41\\
        +Seg+Attr   & 75\% & 19.00 & 35.76\\\hline
        +Seg+Attr+Hier  & 25\% & 17.02 & 33.99\\
        +Seg+Attr+Hier  & 50\% & 18.26 & 37.26\\
        +Seg+Attr+Hier  & 75\% & 19.30 & 37.86\\\hline
        +BBox+Attr+Hier & 25\% & 16.06 & 31.75\\
        +BBox+Attr+Hier & 50\% & 17.21 & 33.46\\
        +BBox+Attr+Hier & 75\% & 18.77 & 35.84\\\hline
        +Attr+Seg+Hier  & 25\% & 16.59 & 32.15\\
        +Attr+Seg+Hier  & 50\% & 17.92 & 34.33\\
        +Attr+Seg+Hier  & 75\% & 18.52 & 36.97\\\hline
    \end{tabular}}
    \vspace{.3em}
  \caption{The effect of diverse supervision under the \FewC~setting. Table for Figure~\ref{sup::fig::multi_sup}, left column.}
    \label{sup::tab::fig1left}
\end{table}

\begin{table}[htbp]
    \centering
    \resizebox{0.9\linewidth}{!}{
    \begin{tabular}{cc|cc}
    \hline
        Model & Portion of Images & Top-1 & Top-5 \\
        \hline\hline
        Baseline & 25\% & 12.64 & 26.13\\
        Baseline & 50\% & 13.96 & 29.21\\
        Baseline & 75\% & 14.39 & 30.66\\\hline
        +Seg    & 25\% & 12.54 & 26.69\\
        +Seg    & 50\% & 14.81 & 31.40\\
        +Seg    & 75\% & 17.07 & 35.50\\\hline
        +Seg+Attr   & 25\% & 12.28 & 26.33\\
        +Seg+Attr   & 50\% & 16.57 & 33.49\\
        +Seg+Attr   & 75\% & 17.94 & 35.90\\\hline
        +Seg+Attr+Hier & 25\% & 13.05 & 27.80\\
        +Seg+Attr+Hier & 50\% & 16.52 & 33.07\\
        +Seg+Attr+Hier & 75\% & 18.03 & 35.25\\\hline
        +BBox+Attr+Hier & 25\% & 12.84 & 27.71\\
        +BBox+Attr+Hier & 50\% & 15.81 & 32.57\\
        +BBox+Attr+Hier & 75\% & 17.42 & 34.55\\\hline
        +Attr+Seg+Hier  & 25\% & 12.76 & 27.59\\
        +Attr+Seg+Hier  & 50\% & 16.11 & 32.52\\
        +Attr+Seg+Hier  & 75\% & 16.96 & 35.12\\\hline
    \end{tabular}}
    \vspace{.3em}
  \caption{The effect of diverse supervision under the \FewI~setting. Table for Figure~\ref{sup::fig::multi_sup}, right column.}
    \label{sup::tab::fig1right}
\end{table}

\begin{table}[htbp]
    \centering
    \resizebox{0.9\linewidth}{!}{
    \begin{tabular}{cc|cc}
    \hline
        Setting & Portion & Top-1 & Top-5 \\
        \hline\hline
        Scarce-Class          & 25\% & 15.08 & 30.96\\
        Scarce-Image           & 25\% & 12.64 & 26.13\\
        Scarce-Class-Adjust   & 25\% & 12.47 & 27.26\\\hline
        Scarce-Class          & 50\% & 15.36 & 31.20\\
        Scarce-Image           & 50\% & 13.96 & 29.21\\
        Scarce-Class-Adjust   & 50\% & 13.37 & 29.00\\\hline
        Scarce-Class          & 75\% & 14.96 & 32.50\\
        Scarce-Image           & 75\% & 14.39 & 30.66\\
        Scarce-Class-Adjust   & 75\% & 14.93 & 31.70\\\hline
    \end{tabular}}
    \vspace{.3em}
  \caption{Performance comparison of the Scarce-Class, Scarce-Image and Scarce-Class-Adjust settings. Table for Figure~\ref{sup::fig::adjust}.}
    \label{sup::tab::fig2}
\end{table}

\begin{table}[htbp]
    \centering
    \resizebox{0.9\linewidth}{!}{
    \begin{tabular}{cc|cc}
    \hline
        Model & Portion of Classes & Top-1 & Top-5 \\
        \hline\hline
        Baseline-Full Classes & 100\% & 17.41 & 33.81\\\hline
        Baseline & 25\% & 14.51	& 29.81\\
        Baseline & 50\% & 14.63	& 30.21\\
        Baseline & 75\% & 14.81	& 31.38\\\hline
        +Seg    & 25\% & 16.08 & 33.05\\
        +Seg    & 50\% & 17.57 & 34.81\\
        +Seg    & 75\% & 16.95 & 34.04\\\hline
        +Seg+Attr   & 25\% & 16.42 & 32.94\\
        +Seg+Attr   & 50\% & 18.36 & 35.03\\
        +Seg+Attr   & 75\% & 18.54 & 35.98\\\hline
        +Seg+Attr+Hier  & 25\% & 16.59 & 33.91\\
        +Seg+Attr+Hier  & 50\% & 17.07 & 34.41\\
        +Seg+Attr+Hier  & 75\% & 18.65 & 36.57\\\hline
    \end{tabular}}
    \vspace{.3em}
    \caption{Performance of Prototypical Network with diverse supervision. Table for Figure~\ref{sup::fig::proto}, left column.}
    \label{sup::tab::fig4l}
\end{table}

\begin{table}[htbp]
    \centering
    \resizebox{0.9\linewidth}{!}{
    \begin{tabular}{cc|cc}
    \hline
        Model & Portion of Images & Top-1 & Top-5 \\
        \hline\hline
        Baseline-Full Images & 100\% & 17.41 & 33.81\\\hline
        Baseline & 25\% & 11.43	& 25.69\\
        Baseline & 50\% & 13.10	& 28.38\\
        Baseline & 75\% & 13.77	& 29.71\\\hline
        +Seg    & 25\% & 12.42 & 26.85\\
        +Seg    & 50\% & 14.33 & 30.16\\
        +Seg    & 75\% & 16.44 & 32.90\\\hline
        +Seg+Attr   & 25\% & 12.62 & 26.66\\
        +Seg+Attr   & 50\% & 15.93 & 32.82\\
        +Seg+Attr   & 75\% & 16.79 & 33.50\\\hline
        +Seg+Attr+Hier  & 25\% & 13.24 & 27.80\\
        +Seg+Attr+Hier  & 50\% & 15.92 & 32.27\\
        +Seg+Attr+Hier  & 75\% & 17.66 & 34.59\\\hline
    \end{tabular}}
    \vspace{.3em}
    \caption{Performance of Prototypical Network with diverse supervision. Table for Figure~\ref{sup::fig::proto}, right column.}
    \label{sup::tab::fig4r}
\end{table}

\end{document}